\RequirePackage{fix-cm}
\RequirePackage{rotating}
\documentclass[twocolumn]{svjour3}          
\smartqed  

\usepackage[utf8]{inputenc} 
\usepackage[T1]{fontenc}    
\usepackage{url}            
\usepackage{booktabs}       
\usepackage{amsfonts}       
\usepackage{nicefrac}       
\usepackage{microtype}      
\usepackage[table]{xcolor}  
\usepackage{epsfig}
\usepackage{graphicx}
\usepackage{amsmath}
\usepackage{amssymb}
\usepackage{multirow}
\usepackage{color, colortbl}
\usepackage{subcaption}
\usepackage{subfloat}
\usepackage{makecell}
\usepackage{tabularx}
\usepackage[export]{adjustbox}
\usepackage{threeparttable}
\usepackage{pifont}
\usepackage[misc]{ifsym} 
\usepackage{arydshln}
\usepackage{afterpage}
\usepackage[ruled,linesnumbered]{algorithm2e}
\usepackage{paralist}
\usepackage[accsupp]{axessibility}

\definecolor{COLOR_CSID}{HTML}{e0f5ff}
\definecolor{COLOR_NEAROOD}{HTML}{ffefe0}
\definecolor{COLOR_FAROOD}{HTML}{ffdebf}
\definecolor{COLOR_MEAN}{HTML}{f0f0f0}

\newcommand{\hutnew}{\textcolor{black}}
\newcommand{\hut}{\textcolor{black}}

\newcommand{\yrexten}{\textcolor{black}}

\usepackage{xspace}
\usepackage{comment}
\usepackage{bm}
\usepackage{wrapfig}
\usepackage[pagebackref=false, breaklinks=true,letterpaper=true, colorlinks,citecolor=citecolor, linkcolor=linkcolor, bookmarks=false]{hyperref}
\definecolor{citecolor}{HTML}{0071BC}
\definecolor{linkcolor}{HTML}{ED1C24}

\makeatletter
\renewcommand\paragraph{
  \@startsection{paragraph} 
  {4} 
  {\z@} 
  {.5em \@plus1ex \@minus.2ex} 
  {-1.5em} 
  {\normalfont\normalsize\bfseries} 
}
\DeclareRobustCommand\onedot{\futurelet\@let@token\@onedot}
\def\@onedot{\ifx\@let@token.\else.\null\fi\xspace}

\makeatother

\makeatletter
\def\makeheadbox{\relax}
\makeatother
\begin{document}
\sloppy 

\title{IAR2: Improving Autoregressive Visual Generation with Semantic-Detail Associated Token Prediction}
\titlerunning{IAR2} 
\author{
        Ran Yi$^{1,\dagger}$ \and
        Teng Hu$^{1}$ \and
        Zihan Su$^1$ \and
        Jiangning Zhang$^2$ \and
        Lizhuang Ma$^1$
}


\institute{
    $\dagger$ Corresponding author. \\
  1 Shanghai Jiao Tong University. \\
    2 Zhejiang University. \\
}

\date{}
\maketitle

\begin{abstract}
Autoregressive models have recently emerged as a powerful paradigm for visual content creation, yet they often overlook the intrinsic structural properties of visual data. Our prior work, IAR, initiated a direction to address this by reorganizing the visual codebook based on embedding similarity, thereby improving generation robustness. However, this approach is constrained by the rigidity of pre-trained codebooks and the inaccuracies of hard, uniform clustering. To overcome these limitations, we propose IAR2, an advanced autoregressive framework that enables a \yrexten{hierarchical} semantic-detail synthesis process. At the core of IAR2 is a novel \textbf{Semantic-Detail Associated Dual Codebook}, which decouples image representations into a semantic codebook for global semantic information and a detail codebook for fine-grained refinements. This design expands the quantization capacity from a linear to a polynomial scale, significantly enhancing expressiveness. To accommodate this dual representation, we \yrexten{propose a \textbf{Semantic-Detail Autoregressive Prediction} scheme coupled with} a \textbf{Local-Context Enhanced Autoregressive Head}, \yrexten{which} performs hierarchical prediction (first the semantic token, then the detail token) while leveraging a local \yrexten{context} window to \yrexten{enhance} spatial coherence. Furthermore, for conditional generation, we introduce a \textbf{Progressive Attention-Guided Adaptive CFG} mechanism that dynamically modulates the guidance scale for each token based on its relevance to the condition \yrexten{and its temporal position in the generation sequence}, improving \yrexten{conditional alignment} without sacrificing realism. Extensive experiments demonstrate that IAR2 sets a new state-of-the-art for autoregressive image generation, achieving a Fréchet Inception Distance (FID) of \textbf{1.50} on ImageNet 256$\times$256. Our model not only surpasses previous methods in performance but also demonstrates superior computational efficiency, highlighting the effectiveness of our structured, coarse-to-fine generation strategy. Code is available at \url{https://github.com/sjtuplayer/IAR2}.

\end{abstract}

\section{Introduction}
\label{sec:intro}

Distinct from diffusion-based~\cite{ddpm} or GAN-based~\cite{goodfellow2020gan} paradigms, which directly operate on the continuous data space, autoregressive and masked image modeling (MIM) frameworks~\cite{sun2024llamagen,VAR,maskgit,meissonic} introduce an additional tokenization step that converts raw images into discrete\yrexten{-valued token} sequences. The subsequent generation process is then formulated as sequence modeling, where autoregressive methods adopt the GPT-style ``next-token prediction" paradigm~\cite{gpt}, while MIM approaches follow the masked-prediction training scheme \yrexten{similar to} BERT~\cite{bert}. Despite their inspiration from natural language modeling, these methods are often directly transplanted to the visual domain without fully accounting for the inherent structural differences between images and text.

\begin{figure}[t]
\centering
\includegraphics[width=0.48\textwidth]{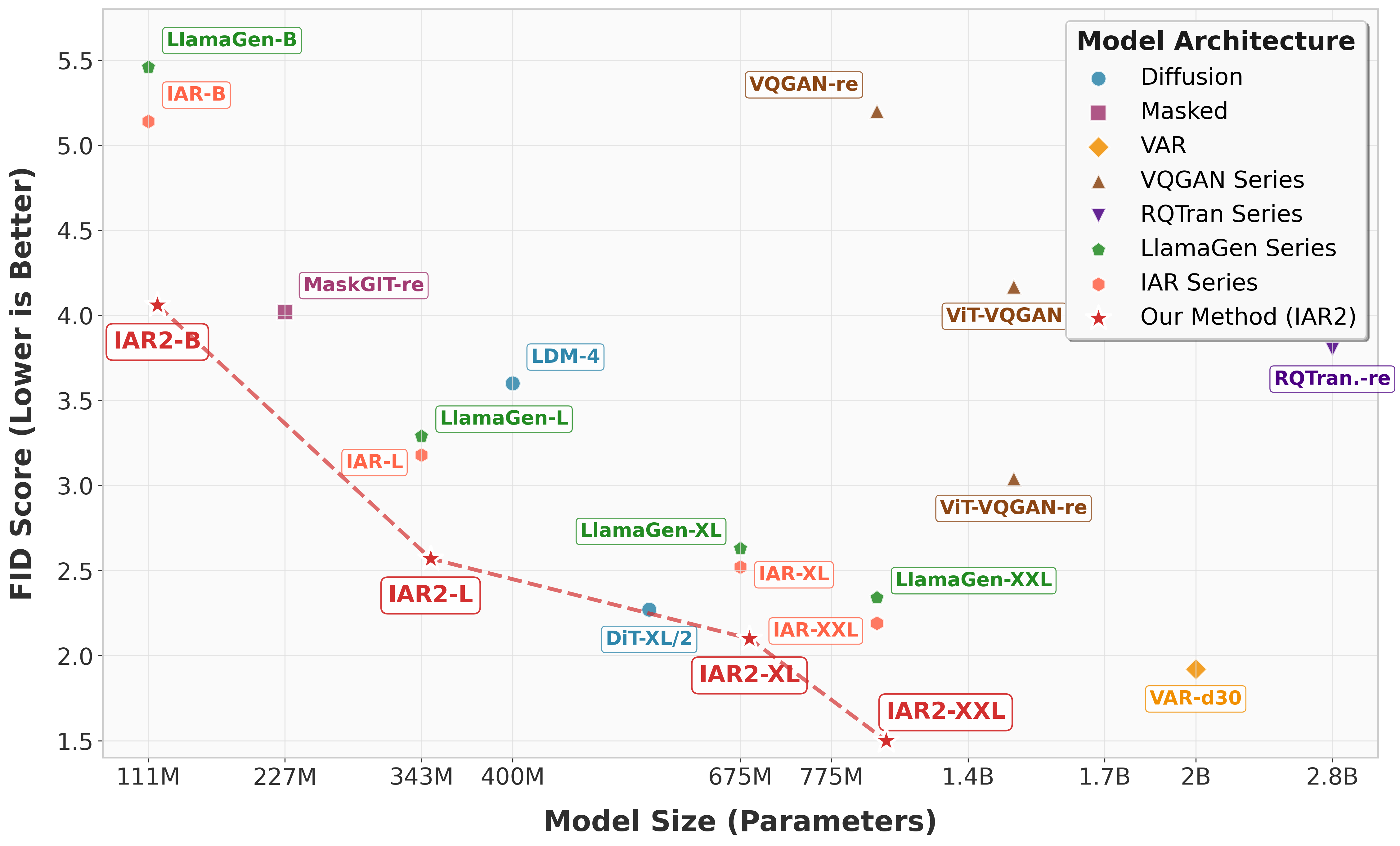}
\caption{Performance Comparison with the state-of-the-art methods on ImageNet. Our model can always achieve the best FID under the same model parameters. Moreover, our IAR2 also achieves the best FID (\textbf{FID=1.50}) across different model and model sizes. }
\label{fig:teaser}
\end{figure}

To better exploit the intrinsic characteristics of visual data, our recent work, \yrexten{Improved} AutoRegressive \yrexten{Visual} Generation (\textbf{IAR}), which is published in CVPR 2025~\cite{hu2025iar}, investigates the relationship between image embeddings and the resulting visual outputs. We observe that embeddings with high similarity typically correspond to images with similar visual content, suggesting that the underlying semantics of an image remain largely stable when represented by closely related \yrexten{image} embeddings. Motivated by this property, IAR \yrexten{proposes a novel \textbf{codebook rearrangement} strategy that} reorganizes the pretrained visual codebook by clustering embeddings into groups of equal size, where tokens within the same cluster exhibit strong similarity. Building upon the reordered codebook, we further introduce a \textbf{cluster-oriented cross-entropy loss}, which encourages the model to first predict the correct cluster before identifying the exact token within it. Since the number of clusters is substantially smaller than the full vocabulary size, the prediction task becomes easier, and even if the model mispredicts the exact token, \yrexten{the token is very likely located in the target cluster, so that} the generated image still remains highly consistent with the ground truth. This design significantly improves the robustness of autoregressive visual generation.  

However, the clustering-based reordering of a pretrained codebook in IAR 
\yrexten{presents certain limitations.}
Directly partitioning a high-dimensional codebook into equal-sized clusters may result in inaccurate groupings, where semantically distinct tokens are erroneously merged, or clusters with inherently different sizes are forcibly divided uniformly. Such inaccuracies can adversely affect the overall performance of the generative model.

To address these issues, we propose \textbf{IAR2}, an advanced autoregressive image generation framework designed to enable a semantic-detail synthesis process and overcome the constraints \yrexten{encountered} by a single codebook. 
\yrexten{We first analyze and find} that single-codebook \yrexten{AR generation} approaches suffer from a trade-off between reconstruction fidelity and \yrexten{generation quality. The observation motivates us to propose} a \textbf{Semantic-Detail Associated Dual Codebook} which decouples image representation into a \emph{Semantic Codebook} that captures global \yrexten{semantic} information and a \emph{Detail Codebook} that focuses on fine-grained local refinements. 
Given an image embedding, the model first retrieves its semantic representation from the semantic codebook \yrexten{(size $n_1$)}, and then encodes the residual information into the detail codebook \yrexten{(size $n_2$)}, thus enabling a two-level quantization. This design expands the effective representational capacity from linear to polynomial scale ($n_1 \times n_2$), substantially enhancing expressiveness compared to conventional single-codebook quantization.

To make autoregressive modeling compatible with the dual-codebook representation, we propose a \textbf{Semantic-Detail Autoregressive Prediction} scheme coupled with a \textbf{Local-Context Enhanced Autoregressive Head}. 
\yrexten{With the semantic-detail dual-codebook, the AR model needs to predict a pair of semantic index and detail index for each patch. Observing that the semantic and detail representations are dependent, we perform a hierarchical prediction of semantic and detail tokens using an AR head.}
At each generation step, the \yrexten{AR} head predicts the semantic token first (coarse prediction), followed by \yrexten{predicting} the detail token (fine prediction) \yrexten{conditioned on the predicted semantic token}. 
In addition, leveraging the \yrexten{inherent} spatial locality of images, our autoregressive head incorporates \yrexten{local contextual cues}, conditioning token prediction on embeddings \yrexten{within a local perception window}. This \yrexten{local context enhancement design} effectively models local dependencies and strengthens spatial coherence, resulting in visually consistent generations.

\hut{Finally, we observe that \yrexten{in conditional generation,} the optimal \yrexten{CFG} guidance \yrexten{scale} is not static. Spatially, the relevance of conditional information varies across an image: salient subjects demand stronger guidance for alignment, while backgrounds benefit from weaker constraints to preserve realism.} Sequentially, as the model generates more \yrexten{patches} of the image, its internal context strengthens, altering the optimal balance between adhering to the external condition and maintaining internal coherence. To address these dual dynamics, we propose \textbf{Progressive Attention-Guided CFG}. Our mechanism modulates the guidance scale for each token based on both its spatial relevance, measured by attention score, and its temporal position in the generation sequence via a progressive schedule. This ensures \yrexten{that the} guidance is applied precisely where and when it is most needed, significantly improving conditional alignment without sacrificing overall image quality.

Extensive experiments demonstrate that \textbf{IAR2} substantially advances the state-of-the-art in autoregressive image generation. Notably, it reduces the FID of the 100M-parameter LlamaGen model from \textbf{6.09} to \textbf{4.80}, and achieves an FID of \textbf{1.50} with the 1.5B-parameter IAR2-XXL, outperforming the 2B-parameter VAR (FID 1.92) trained on 256 GPUs, while IAR2 attains superior performance using only 32 GPUs. \hutnew{Beyond visual quality, our framework significantly accelerates the optimization process, yielding an approximate 62\% increase in training efficiency.} These results highlight the efficacy and strong \emph{scaling-up} capability of our approach. \hutnew{Compared to our preliminary conference version (IAR), this extended work introduces a fundamentally upgraded architecture with the following newly developed contributions:}
\begin{itemize}
    \item We propose a \textbf{Semantic-Detail Associated Dual Codebook Quantization} that decomposes image representations into a semantic codebook for global \yrexten{semantics} and a detail codebook for local refinements, expanding \yrexten{representational} capacity from linear to polynomial scale for more expressive coarse-to-fine generation.
    
    \item We \yrexten{design} a \textbf{Local-Context Enhanced Autoregressive Head} tailored to the dual-codebook \yrexten{AR generation}. It performs hierarchical prediction (semantic then detail token), and incorporates a local perception window to condition each prediction on nearby spatial information, \yrexten{thereby} significantly improving local coherence of generated images.
    
    \item  \hut{We propose a \textbf{Progressive Attention-Guided CFG} that dynamically modulates the guidance scale for each token based on its spatial relevance and sequential progress. It leverages attention mechanism to concentrate guidance on salient regions and employs a progressive schedule to intensify its strength as generation proceeds, thereby improving condition alignment while preserving overall image quality.}
\end{itemize}

\section{Related Work}

\subsection{Visual Tokenizers}
A core component of discrete visual generation is the tokenizer, which maps continuous images into compact sequences of discrete tokens. Single-codebook quantizers such as VQ-VAE\cite{van2017vqvae}, VQGAN\cite{esser2021vqgan}, and ViT-VQGAN\cite{yu2021ViT-VQGAN} employ a learnable codebook to quantize feature vectors. While these methods enable effective compression, their representational capacity is fundamentally constrained by the size of a single codebook.

To improve quantization accuracy and increase the diversity of discrete representations, multi-codebook quantization has been proposed. RQ-VAE\cite{lee2022rqvae} adopts a residual quantization strategy, encoding the residual between the target vector and its reconstruction with additional codebooks, thereby progressively enhancing fidelity. FQGAN\cite{bai2024fqgan}, UniTok\cite{ma2026unitok}, and TokenFlow\cite{qu2025tokenflow} decompose feature channels and quantize them with multiple codebooks, leading to a combinatorial increase in representational capacity. MAGVIT-v2\cite{yu2023magvit,luo2024Open-magvit2} further introduces a radix-based quantization scheme that eliminates explicit codebooks and achieves lookup-free quantization. DualToken\cite{song2025dualtoken} integrates semantic and pixel-level information across different layers of a vision encoder, achieving state-of-the-art reconstruction performance, though it does not explore generative modeling.

\subsection{Continuous-Valued Visual Generation}
Early research in visual synthesis was dominated by continuous-valued generative models. Generative Adversarial Networks (GANs)\cite{goodfellow2020gan,mirza2014cgan,radford2015dcgan,salimans2016improvedgans,isola2017Pix2Pix,zhu2017CycleGAN} pioneered adversarial training between a generator and discriminator, leading to high-fidelity image synthesis, with subsequent advances such as StyleGAN\cite{karras2019stylegan} pushing visual realism further. However, GANs often suffer from unstable training and mode collapse. More recently, diffusion models\cite{ho2020ddpm,nichol2021iddpm,song2020ddim,dhariwal2021adm} have emerged as the prevailing paradigm, generating high-quality and diverse samples through iterative denoising. Large-scale extensions such as Imagen\cite{saharia2022Imagen} and Stable Diffusion\cite{rombach2022stable} have advanced text-to-image generation to new levels. Despite these successes, both GANs and diffusion models inherently operate in the continuous domain, making them less compatible with discrete sequence modeling frameworks inspired by large language models.

\subsection{Discrete-Valued Visual Generation} 
To align visual synthesis with the principles of language modeling, recent approaches discretize images into token sequences for generation in the discrete\yrexten{-valued} domain.

\noindent\textbf{Single-Codebook Autoregressive Models.}
Autoregressive (AR) image generation follows the “next-token prediction” paradigm of GPT \cite {gpt,gpt4}, where each image token is sequentially predicted conditioned on previously generated ones. Early works such as VQGAN+Transformer \cite {esser2021vqgan}, DALL·E \cite {ramesh2021DALLE}, and Parti \cite {yu2022Parti} quantize images into a single codebook and then model the generation process with Transformers. More recent advances include LlamaGen \cite {sun2024llamagen}, which leverages the LLaMA framework \cite {llama} to enhance semantic modeling, and VAR \cite {VAR}, which introduces a progressive multi-scale generation pipeline.

Parallel to autoregressive modeling, another line of research is Masked Image Modeling (MIM), which follows the “mask-and-predict” strategy inspired by BERT~\cite{bert}. By reconstructing masked regions in parallel, MIM can improve decoding efficiency. Representative works such as MaskGIT \cite{chang2022maskgit}, MagViT \cite{yu2023magvit}, and MUSE \cite{chang2023muse} enable partially parallel decoding and achieve faster generation compared to fully autoregressive models. 

These works establish the viability of modeling images as discrete token sequences. However, they are fundamentally constrained by the reliance on a single codebook: with a small codebook, reconstruction quality is poor and thus the upper bound of generation fidelity is limited; with a large codebook, reconstruction improves, but the model faces substantially higher difficulty in predicting tokens from an enlarged candidate space.

\noindent\textbf{Multi-Codebook Autoregressive Models.}
\hut{To overcome the single-codebook bottleneck, multi-codebook AR models have been developed.
Recent methods like Dual-Token~\cite{song2025dualtoken} and FQGAN~\cite{bai2024fqgan} often follow the MAGVIT-v2~\cite{yu2023MAGVIT-v2} paradigm, where multiple codebooks 
\yrexten{operate in parallel during quantization, lacking the semantic association needed to}
form a cohesive, hierarchical representation of the content. 
Although TokenFlow~\cite{qu2025tokenflow} utilizes separate semantic and pixel-level codebooks, its design imposes a one-to-one mapping between them, which fundamentally limits its representational capacity to a linear scale and hinders its generative potential. \yrexten{And its generative process still relies on a single-codebook prediction.}}

In summary, continuous-valued models (GANs, diffusion) have established strong baselines for high-quality generation, while discrete-valued frameworks, particularly multi-codebook AR models, offer a promising direction for bridging visual generation with the scaling properties of large language models. \hut{However, a key limitation of existing multi-codebook frameworks is their failure to model the hierarchical relationship between tokens, \yrexten{either} treating 
\yrexten{multiple codebooks
as uncorrelated or enforcing an overly strict one-to-one mapping.}} 
This can lead to inefficient modeling and a weaker enforcement of semantic consistency. 
Our work advances this line of research by proposing a \yrexten{semantic-detail associated} dual-codebook autoregressive framework that explicitly models the interplay between semantic and detail representations \yrexten{and performs hierarchical prediction}, thereby strengthening modeling at both levels.

\section{The IAR Approach}
\label{sec:iar}

\begin{figure}[t]
\centering
\includegraphics[width=0.48\textwidth]{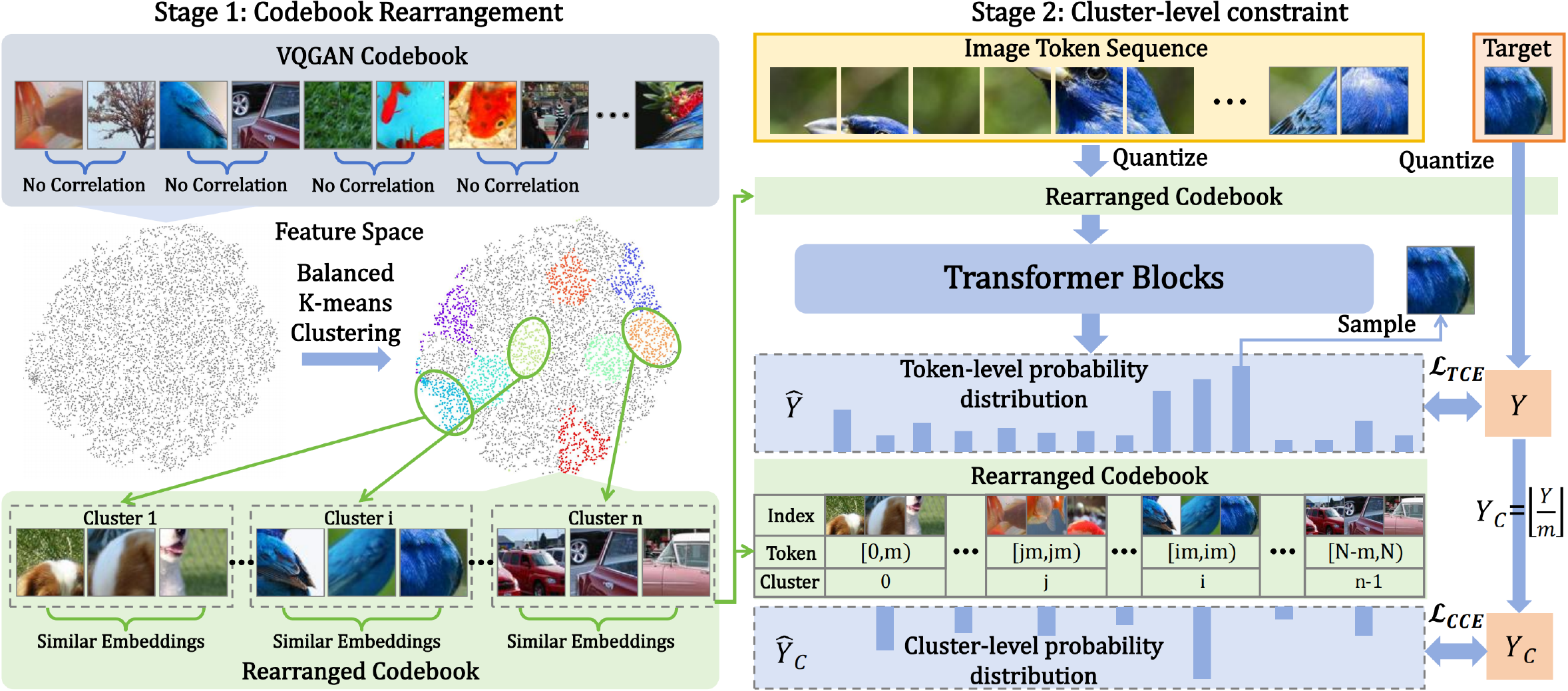}
\caption{The IAR Framework: IAR begins by rearranging its codebook to group semantically similar image embeddings into distinct clusters. Subsequently, during the training of the autoregressive model, IAR introduces a cluster-level constraint. This constraint guides the model to predict the correct cluster index for a given image, ensuring that the generated embedding is close to the target. This approach significantly enhances the robustness and overall performance of the AR model. }
\label{fig:iar main framework}
\vspace{-0.1in}
\end{figure}

In conventional text generation, predicted indices directly map to words. 
\yrexten{In contrast,} image generation requires an additional step: mapping indices to embeddings that are subsequently decoded into images. 
We observe that nearby embeddings usually represent semantically and visually similar patches, such that replacing a patch \yrexten{embedding} with a close embedding yields nearly identical \yrexten{decoded images}.

Motivated by this, we introduce \textbf{IAR}, a framework that exploits the structure of the embedding space to enhance LLM-based image generation. \yrexten{As shown in Fig.~\ref{fig:iar main framework},} IAR comprises two \yrexten{components}: (1) \emph{Codebook rearrangement}, which employs balanced K-means to cluster embeddings into equally sized groups of high intra-cluster similarity, ensuring that cluster-level accuracy compensates for token-level errors; and (2) \emph{Cluster-oriented cross-entropy}, which relaxes supervision from exact tokens to clusters, thus \yrexten{guiding} predictions 
\yrexten{towards correct cluster and semantics.}
Together, these strategies make IAR robust to token errors while improving training efficiency and ensuring stable, high-quality image generation.

\subsection{Analysis on Image Embedding Similarity}
The design of IAR is motivated by a fundamental property of visual tokenizers: embeddings that are close in the latent space typically encode visually similar content. This implies that replacing a token embedding with a nearby one in the latent space results in a decoded image that is nearly identical to the original in both semantics and appearance.

To verify this property, we conducted experiments on the VQGAN~\cite{vqgan} codebook following a structured workflow. First, input images were tokenized into discrete embeddings using the VQGAN tokenizer. Next, we progressively replaced these original embeddings with alternatives at varying ``code distances" (\yrexten{euclidean distances between two embeddings in the latent space}), and decoded the modified embeddings back into images. Finally, we quantified the similarity between the reconstructed images and their originals using two metrics: mean squared error (MSE) and Learned Perceptual Image Patch Similarity (LPIPS)~\cite{lpips}.

\begin{figure}[t]
\centering
\includegraphics[width=0.48\textwidth]{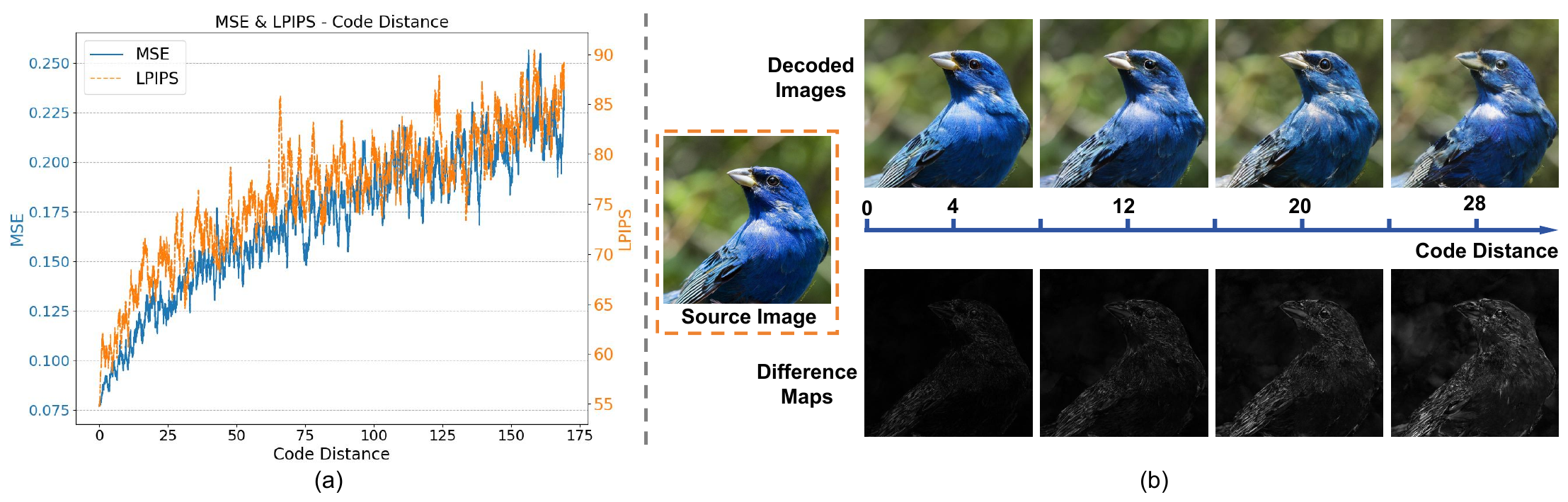}
\caption{(a) \yrexten{The} MSE and LPIPS between the source image and the reconstructed image under different code distances. (b) Visualization of decoded images at varying code distances.}
\label{fig:iar code distance}
\vspace{-0.1in}
\end{figure}

As shown in Fig.~\ref{fig:iar code distance}, the experimental results reveal two key trends. First, as the code distance between the original and replacement embeddings increases, \yrexten{the} discrepancies between the reconstructed and original images gradually grow larger. Second, at small code distances (e.g., distances \textless 12), these differences are negligible, such that the decoded images remain visually indistinguishable from the originals. This finding demonstrates the inherent robustness of the embedding space: even when the model predicts an incorrect token index, if the corresponding embedding is close to the ground-truth embedding in the latent space, the decoded image retains semantic fidelity to the target \yrexten{image}. Leveraging this property, IAR integrates codebook rearrangement and cluster-level loss to significantly enhance the stability and quality of LLM-driven image generation.
\subsection{Codebook Rearrangement}
\label{ssec:codebook rearrange}
The codebook learned by VQGAN\cite{vqgan} is typically unordered: adjacent indices often correspond to semantically unrelated embeddings, making index proximity uninformative. We address this with \textbf{Codebook Rearrangement}, which reorders embeddings so that 
\yrexten{adjacent embeddings in codebook exhibit high similarity.}

Formally, given the codebook $\mathcal{Z}=\{z_i\}_{i=1}^N$, the goal is to find a mapping $M(\cdot)$ that minimizes the distance between consecutive embeddings:
\begin{equation}
    \begin{aligned}
        M=\arg\min\limits_{M} \sum_{i=1}^{N-1} \|z_{M(i)},z_{M(i+1)}\|.
    \end{aligned}
\end{equation}
This \yrexten{optimization can be reduced to} Hamiltonian-path \yrexten{problem, which} is NP-hard.
Thus, we \yrexten{relax this problem to an easier one and solve} 
it via clustering.

\yrexten{To relax the problem into a solvable one,}
instead of enforcing index adjacency on a global scale, we require local similarity within clusters. Specifically, the codebook is partitioned into $n$ clusters of equal size $m=\frac{N}{n}$. In the rearranged codebook, embeddings of cluster $j$ occupy indices  $[jm, (j+1)m)$. This ensures that intra-cluster adjacency reflects semantic similarity while keeping the problem tractable.

We adopt a balanced K-means clustering algorithm to construct clusters \yrexten{and rearrange codebook}. 
This algorithm ensures both high intra-cluster similarity and uniform cluster size. Starting from randomly initialized centers $\{c_j\}_{j=1}^n$, embeddings are iteratively assigned to the nearest available cluster (up to size $m$), and centers are updated as the mean of assigned embeddings. This process converges to $n$ balanced clusters, yielding a reordered codebook where adjacent indices correspond to semantically similar embeddings.

\subsection{Cluster-\yrexten{oriented} Visual Generation}  

Existing LLM-based visual generation models \cite{sun2024llamagen} are trained with Token-\yrexten{oriented} Cross-entropy \yrexten{loss} (TCE):  
\begin{equation}
\mathcal{L}_{TCE} = -\sum_{i=1}^N Y_i \log \hat{Y}_i,
\tag{4}
\end{equation}
where $Y$ and $\hat{Y}$ denote the one-hot ground-truth and predicted distributions over $N$ tokens.  
However, TCE penalizes all incorrect tokens equally, ignoring latent-space similarity: predicting a \yrexten{highly similar} embedding often yields nearly identical decoded images.  
With the rearranged codebook (Sec.~\ref{ssec:codebook rearrange}), embeddings within a cluster are semantically consistent. 
\yrexten{This shifts the critical prediction task from identifying the exact token to predicting the correct cluster, which largely determines the semantics of the generated images.}
This observation inspires a \yrexten{two-level} supervision strategy: first, ensure the correct prediction of clusters, and then refine the \yrexten{prediction} of \yrexten{specific} tokens.

{\bf Cluster-oriented Cross-entropy Loss.} We define the cluster label of token $y$ as $y_c=\lfloor \frac{y}{m} \rfloor$, where each cluster contains $m=\frac{N}{n}$ tokens.  
The predicted cluster distribution $\hat{Y}_{C} \in \mathcal{R}^n (\sum \hat{Y}_{C,i}=1)$ is obtained by summing token probabilities within each cluster:  
\begin{equation}
\hat{Y}_{C,j} = \frac{\sum_{i=jm}^{(j+1)m-1} \exp(\hat{Y}_i)}{\sum_{i=1}^N \exp(\hat{Y}_i)}, 
\quad j = 1, \dots, n.
\end{equation}
The Cluster-level Cross-entropy (CCE) loss is then \yrexten{formulated as}:  
\begin{equation}
\mathcal{L}_{CCE} = -\sum_{j=1}^n Y_{C,j} \log \hat{Y}_{C,j},
\end{equation}
\yrexten{where $Y_C$ is the one-hot vector spanned by cluster label $y_c$. CCE loss}
rewards correct cluster prediction even if the exact token is wrong, improving robustness and semantic fidelity.  

{\bf Final Loss Function.} The overall objective combines both levels of supervision:  
\begin{equation}
\mathcal{L}=\mathcal{L}_{TCE}+\lambda\mathcal{L}_{CCE},
\end{equation}
where $\lambda$ balances cluster-level \yrexten{accuracy} with token-level precision.

\section{IAR2}
\label{sec:iar++}

In this section, we propose \textbf{IAR2}, an advanced autoregressive image generation framework designed to overcome the limitations of its predecessor, IAR \yrexten{(Sec.~\ref{sec:iar})}, which relies on a single, pre-trained codebook. 
We begin by analyzing the fundamental trade-off in conventional single-codebook \yrexten{AR generation} approaches: achieving higher reconstruction fidelity necessitates a larger codebook. However, an expanded codebook size exponentially increases the modeling challenge for the generative model. 
\yrexten{This is because, due to the large number of classes, it becomes difficult to predict a correct class label, which} often leads to a degradation in the final generation quality.

This observation motivates our core innovation: a \textbf{Semantic-Detail Associated Dual Codebook} that decouples visual representation into semantic and detail components. Specifically, we employ a compact codebook (with size $n_1$) to capture high-level semantic information, and a significantly larger codebook (with size $n_2$) \yrexten{associated with the former} to encode fine-grained details and textures. 
\yrexten{With the dual codebook, we encode an image into semantic and detail codes, which are then used for autoregressive generation.}
\yrexten{Combining these two codebooks} effectively expands the theoretical representational capacity to $n_1 \times n_2$, far exceeding that of a single-codebook system.

To extend AR models to a dual-codebook (semantic, detail) framework, we propose the \textbf{Semantic-Detail Autoregressive Prediction} scheme. 
It represents each semantic-detail token pair \yrexten{within} a single hidden state \yrexten{to} maintain the original sequence length. 
From this state, \yrexten{for each patch,} the \yrexten{AR} model hierarchically predicts the semantic token first, followed by \yrexten{predicting} the detail token \yrexten{conditioned on the predicted semantic token}. 
This approach leverages the 
\yrexten{manageable size}
of the compact semantic codebook\yrexten{, enabling more accurate semantic prediction from a limited vocabulary. This reliable semantic prediction} ensures that the generated image preserves semantic coherence,
making \yrexten{the generation} robust to minor errors in detail prediction.

However, simply using MLPs to \yrexten{independently} predict semantic and detail tokens 
\yrexten{overlooks the correlation between semantic tokens and detail tokens.}
To address this, we introduce a novel \textbf{Local-Context Enhanced Autoregressive Head} that is specifically designed for sequential, hierarchical token prediction. This AR head operates on the intermediate embeddings produced by the main autoregressive model and, at each decoding step, first predicts the semantic token and then the corresponding detail token, all within the same hidden representation. Importantly, it incorporates \yrexten{local} contextual cues from previously generated tokens within a local window, \yrexten{using them to model} complex visual dependencies 
\yrexten{and enhance local spatial coherence.}
This ensures that hierarchical prediction is both feasible and effective, further strengthening the model's structural coherence and visual fidelity.

Finally, during the inference stage, to overcome the limitations of conventional CFG, namely its \textit{Spatial Uniformity}, which \yrexten{causes} artifacts in the background, and its \textit{Sequential Staticity}, 
\hut{which ignores that the generation process is evolutionary; as more of the image is generated, the model builds a stronger internal context, altering the optimal balance between adhering to the external condition and maintaining internal coherence}, we propose \textbf{Progressive Attention-Guided CFG (PAG-CFG)}. This mechanism uses attention scores to \yrexten{adjust} guidance \yrexten{scale to focus} on semantically relevant regions, and employs a progressive schedule to adapt guidance strength as generation evolves from coarse composition to fine-grained refinement. As a result, the generated images exhibit stronger alignment with conditioning prompts, leading to a significant improvement in overall generation fidelity and quality.

\subsection{Impact of Codebook Size on Reconstruction and Generation Capabilities}

\begin{figure}[t]
\centering
\includegraphics[width=0.48\textwidth]{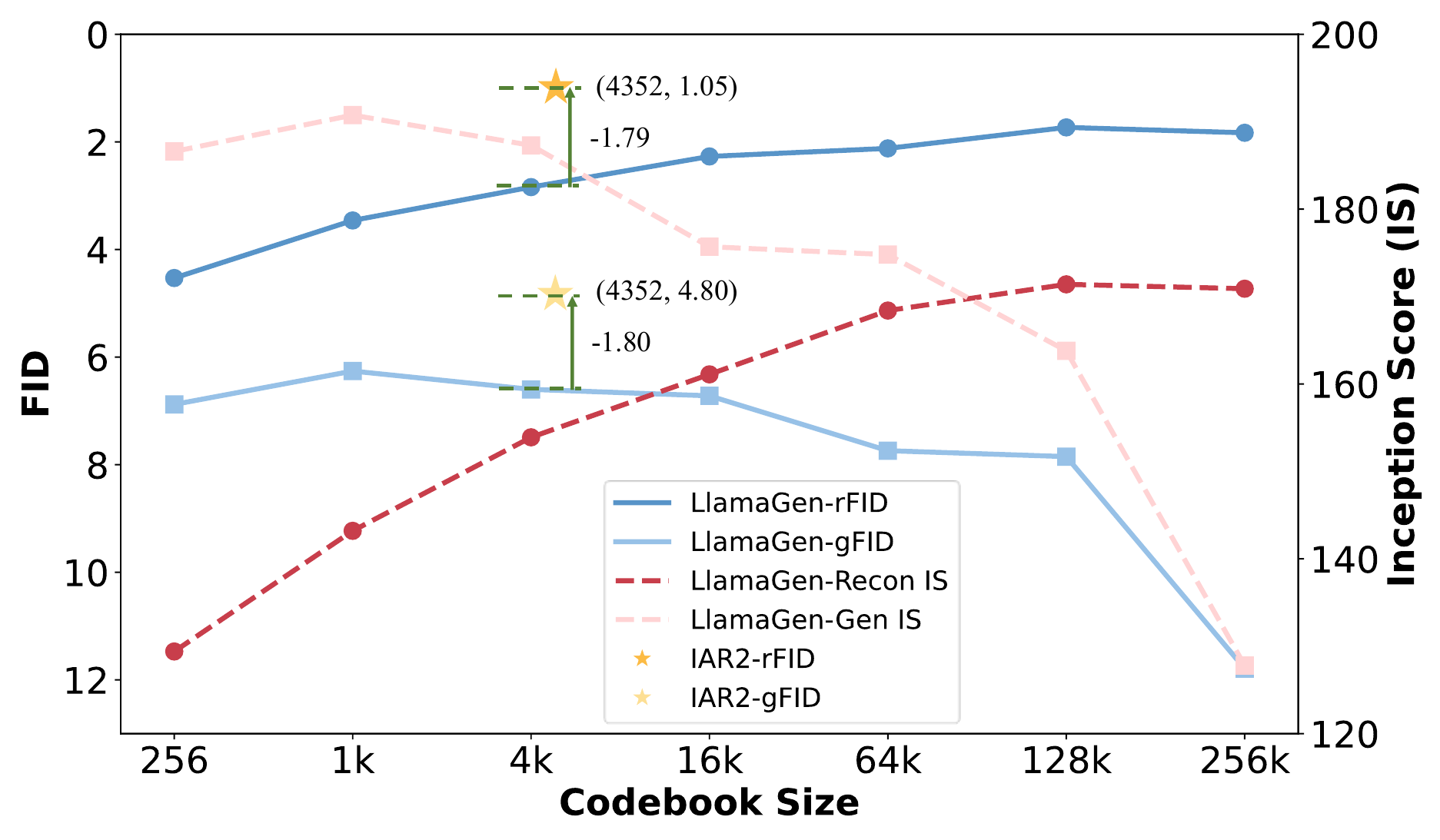}
\caption{Impact of codebook size on reconstruction and generation. While increasing the codebook size enhances reconstruction accuracy, an excessively large codebook complicates the learning task for the generative model, leading to degraded generation quality. In contrast, our semantic-detail associated quantization strategy strikes an effective balance, achieving high fidelity in both reconstruction and generation.}
\label{fig:motivation}
\end{figure}

Unlike GANs and diffusion models, which operate on continuous visual representations, autoregressive (AR) \yrexten{image generation} models necessitate the discretization of images into a sequence of tokens. This is typically achieved through a vector-quantized codebook trained via a framework like VQGAN. However, this quantization process inevitably leads to information loss. Intuitively, a larger codebook size should reduce this loss, allowing the discrete representation to more closely approximate the continuous space. This raises a critical question: \textit{can we indefinitely increase the codebook size to enhance reconstruction fidelity and, consequently, generative quality?}

\textbf{Impact of Codebook Size on Reconstruction Capability.} To investigate this question, we conduct an empirical study to analyze the relationship between codebook size, reconstruction fidelity, and generative performance. We train a series of VQGAN models on the ImageNet dataset~\cite{deng2009imagenet} with seven distinct codebook sizes, from 256\yrexten{, 1k, 4k, ...,} to 256k. First, we evaluate their reconstruction capabilities. As illustrated in Fig.~\ref{fig:motivation}, a clear trend emerges: as the codebook size increases, the reconstruction FID (rFID) consistently decreases\yrexten{, indicating better reconstruction fidelity}. This confirms that a larger vocabulary enhances the codebook's representational power, thereby establishing a higher theoretical upper bound for the quality of the final generated images.

\textbf{Impact of Codebook Size on Generation Capability.} We then examine the impact on the generative task itself. Using each pre-trained VQGAN~\cite{vqgan} as a tokenizer, we train a LlamaGen~\cite{sun2024llamagen} model with 111M parameters to generate images autoregressively from the corresponding discrete tokens. We generate 50,000 samples for each \yrexten{codebook size} configuration, and compute the generative FID (gFID) against the ground truth distribution. The results, also shown in Fig.~\ref{fig:motivation}, reveal a more complex, non-monotonic relationship. Initially, the gFID improves, dropping from 6.4 (with a 256-sized codebook) to an optimal 6.1 (with a 1024-sized codebook). However, as the codebook size continues to expand, the gFID begins to degrade, despite the continuous improvement in reconstruction potential (rFID). This demonstrates that beyond a certain threshold, a larger codebook significantly increases the difficulty of the generative modeling task. \yrexten{This is because, with a larger number of classes, it becomes more difficult for the AR model to predict a correct class label.}
The vast, sparse prediction space poses a formidable challenge for the AR model, leading to a decline in generation quality.

From this analysis, we draw two key conclusions:  \textit{\textbf{(1)} Increasing the codebook size monotonically improves the potential reconstruction fidelity}.  \textit{ \textbf{(2)} There exists an optimal codebook size for generative performance; exceeding this threshold complicates the prediction task to the detriment of generation quality}. 
This fundamental trade-off motivates our proposal to extend the conventional single-codebook paradigm to a dual-codebook architecture. By \yrexten{using two codebooks of size $n_1$ and $n_2$}, we expand the theoretical representational capacity to $n_1 \times n_2$, enhancing reconstruction potential. \yrexten{Meanwhile,} the AR model only needs to predict from two smaller, more tractable \yrexten{codebooks} of size $n_1$ and $n_2$ sequentially\yrexten{, without the need to predict a correct class label from $n_1 \times n_2$ classes}. This approach effectively mitigates the modeling complexity and resolves the aforementioned trade-off.

\begin{figure*}[t]
\centering
\includegraphics[width=1.0\textwidth]{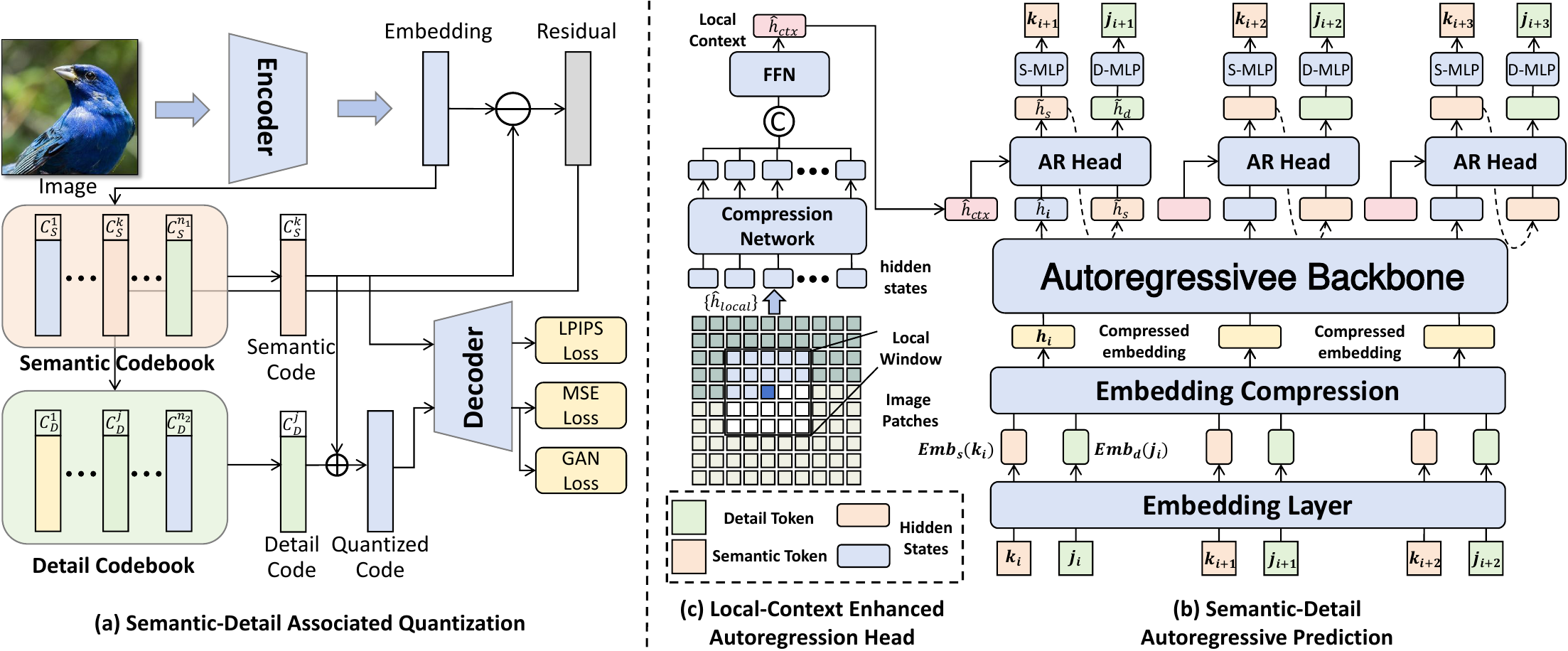}
\caption{IAR2 consists of three main modules: 1) The \textbf{Semantic-Detail Associated Quantization Module} disentangles an input image into two distinct sets of discrete codes: semantic codes for high-level content and detail codes for fine-grained visual information;
2) The \textbf{Semantic-Detail Autoregression Model} processes these token pairs by fusing them into a unified hidden state, which is then fed into an autoregressive backbone to 
\yrexten{obtain global contexts;}
3) The \textbf{Local-Context Enhanced Autoregression Head} \yrexten{performs hierarchical prediction of semantic and detail tokens, and leverages} 
neighboring \yrexten{local} context tokens to enrich the local information, thereby enhancing generation accuracy for both semantic and detail codes.}
\label{fig:main framework of IAR2}
\vspace{-0.1in}
\end{figure*}

\subsection{Semantic-Detail Associated Vector Quantization}

\hut{As established in Section 4.1, conventional single-codebook methods face an inherent trade-off between reconstruction fidelity and generative modeling complexity. To \yrexten{address} this limitation, a dual-codebook architecture presents a promising direction, offering the potential for expanded representational capacity without a proportional increase in modeling difficulty. The key to unlocking this potential, however, lies in the design of the \yrexten{dual} codebook structure and the corresponding generation process.}

\hut{To this end, we draw inspiration from the cluster-oriented principle validated in our prior work, IAR~\cite{hu2025iar}, which demonstrated that separating the prediction of high-level concepts from the refinement of specific details leads to more robust generation. Motivated by this hierarchical strategy, we propose the \textbf{Semantic-Detail Associated Dual Codebook}, a novel vector quantization framework designed to structure the generation process in a semantic-detail manner. As illustrated in Fig.~\ref{fig:main framework of IAR2}~(a), our architecture is composed of two specialized components: (1) A compact \textbf{Semantic Codebook} ($\mathcal{C}_s$), which is designed to capture the high-level semantics, global structure, and essential content of an image patch; (2) A larger \textbf{Detail Codebook} ($\mathcal{C}_d$), which is trained to encode the residual high-frequency information, such as fine textures and local patterns, that remains after the semantic information has been abstracted.
This design facilitates a two-stage, sequential prediction process that mirrors the hierarchical nature of the representation. The autoregressive model first predicts the semantic token, thereby establishing the core visual content. Subsequently, it predicts the detail token to render the fine-grained specifics. This approach transforms the complex task of predicting a single, high-information token into two more manageable and focused sub-problems, ensuring that the generative process is both more robust and capable of leveraging the enhanced expressive power of the dual-codebook system.}

\subsubsection{\yrexten{Semantic-Detail} Vector Quantization}
Our quantization process operates via residual quantization. For a given image patch $I_i$, an encoder network $E$ first maps it to a latent embedding $e_i = E(I_i)$. The quantization then proceeds in two stages:

\begin{enumerate}
    \item \textbf{Semantic Quantization:} We first identify the nearest entry $q_{i,s}$ from the semantic codebook $\mathcal{C}_s = \{c_s^k\}_{k=1}^{n_1}$ to represent the core semantic content:
    \begin{equation}
        q_{i,s} = \underset{q_{i,s}\in \left [1,n_1\right ] }{\arg\min} \| e_i - c_s^{q_{i,s}} \|_2^2.
    \end{equation}
    \item \textbf{Detail Quantization:} We then compute the residual vector $e_{i,res} = e_i - \mathcal{C}_s^{q_{i,s}}$, which captures the \yrexten{fine-grained} information not represented by the semantic code. This residual is subsequently quantized using the detail codebook $\mathcal{C}_d = \{c_d^j\}_{j=1}^{n_2}$:
    \begin{equation}
        q_{i,d} = \underset{q_{i,d} \in \left [1,n_2\right ]}{\arg\min} \| e_{i,res} - c_d^{q_{i,d}} \|_2^2.
    \end{equation}
\end{enumerate}
The final quantized representation for the patch is the sum of the two selected codes, $\hat{e}_i = c_s^{q_{i,s}} + c_d^{q_{i,d}}$, while the discrete representation passed to the generative model is the pair of indices $(k_i, j_i)=(q_{i,s},q_{i,d})$. A decoder $D$ then reconstructs the image patch as $\hat{I}_i = D(\hat{e}_i)$.

\subsubsection{Training Objective for VQ Model}
The training of our \yrexten{semantic-detail associated} codebook is conducted in two stages to ensure that each component learns its designated role effectively. This multi-stage strategy is crucial for disentangling semantic information from fine-grained details.

\textbf{Stage 1: Semantic Codebook Pre-training.}
Initially, we focus exclusively on training the semantic codebook $\mathcal{C}_s$ along with the encoder $E$ and decoder $D$. The goal of this stage is to equip the semantic codebook with the ability to capture the semantic content of images. To achieve this, we optimize the model using only a perceptual loss (LPIPS), which is well-suited for measuring semantic similarity, alongside a standard vector quantization commitment loss~\cite{vqgan}. The parameters of the encoder $E$, decoder $D$, and semantic codebook $\mathcal{C}_s$ are jointly optimized by minimizing the following objective:
\begin{equation}
    \min_{E, D, \mathcal{C}_s} \mathcal{L}_{\text{VQ, Stage 1}} = \mathcal{L}^{s}_{\text{commit}} + \lambda_{perc}\mathcal{L}_{\text{LPIPS}}(I, \hat{I}_s),
    \label{eq:semantic pretrain}
\end{equation}
where $\hat{I}_s = D(c_s^{q_s})$ is the reconstruction based solely on the semantic code. \hut{The semantic commitment loss \yrexten{$\mathcal{L}^{s}_{\text{commit}}$} is defined as follows:
\begin{equation}
\begin{aligned}
    \mathcal{L}^s_{\text{commit}} =& \mathbb{E}_{e \sim E(I)}\left[\lVert \text{sg}[e] - c_s^{q_s} \rVert_2^2 \right] + \\
    &\beta \mathbb{E}_{e \sim E(I)}\left[\lVert e - \text{sg}[c_s^{q_s}] \rVert_2^2 \right],
\end{aligned}
\end{equation}
where $\text{sg}[\cdot]$ denotes the stop-gradient operator, $e$ is the encoded representation, $c_s^{q_s}$ is the quantized semantic code, and $\beta$ controls the strength of the codebook commitment.}

\textbf{Stage 2: Joint Semantic-Detail Training.}
Following the semantic pre-training, we introduce the detail codebook $\mathcal{C}_d$ and proceed to the second stage, where all components ($E$, $D$, $\mathcal{C}_s$, and $\mathcal{C}_d$) are trained jointly. The objective of this stage is to train the detail codebook to capture the high-frequency residual information necessary for high-fidelity reconstruction, while allowing the other components to adapt. The optimization is driven entirely by reconstruction-focused losses: a combination of an L2 loss for pixel-level accuracy, an adversarial (GAN) loss to enhance perceptual realism and sharpness, \yrexten{and a commitment loss}. The full objective for this joint training stage is:
\begin{equation}
\begin{aligned}
    \min_{E, D, \mathcal{C}_s, \mathcal{C}_d} \mathcal{L}_{\text{VQ, Stage 2}} = \mathcal{L}^{sd}_{\text{commit}} &+ \lambda_{rec}\mathcal{L}_{\text{L2}}(I, \hat{I}) \\
    &+ \lambda_{adv}\mathcal{L}_{\text{GAN}}(I, \hat{I}),
    \label{eq:vq,stage2}
\end{aligned}
\end{equation}
\hut{where $\hat{I} = D(c_s^{q_s} + c_d^{q_d})$ is the final reconstruction from both semantic and detail codes. 
The semantic-detail commitment loss \yrexten{$\mathcal{L}^{sd}_{\text{commit}}$} is defined as:
\begin{equation}
\begin{aligned}
    \mathcal{L}^{sd}_{\text{commit}} =& \mathbb{E}_{e \sim E(I)}\left[\lVert \text{sg}[e] - c_s^{q_s} - c_d^{q_d} \rVert_2^2 \right] + \\
    &\beta \mathbb{E}_{e \sim E(I)}\left[\lVert e - \text{sg}[c_s^{q_s} + c_d^{q_d}] \rVert_2^2 \right],
\end{aligned}
\end{equation}
where $\text{sg}[\cdot]$ denotes the stop-gradient operator, $e$ is the encoded representation, $c_s^{q_s}$ and $c_d^{q_d}$ are the quantized semantic and detail codes, respectively, and $\beta$ controls the strength of the codebook commitment. Moreover, to keep the semantic representation ability of the semantic codebook, 
\yrexten{we interleave the joint training (Eq.~\ref{eq:vq,stage2}) with periodic updates to the semantic codebook (Eq.~\ref{eq:semantic pretrain}) at a 2:1 ratio.}} 

\hutnew{Notably, although this dual-codebook optimization and interleaved scheduling introduce a minor increase in per-iteration computational time, the explicit factorization of semantic and detail representations provides a much clearer and more structured learning objective. This significantly accelerates the overall optimization process, leading to a substantially faster convergence rate compared to conventional single-codebook paradigms. Ultimately, this two-stage process first establishes a robust semantic foundation, and then allows both codebooks to collaboratively refine the representation for high-fidelity reconstruction.}

\subsection{Semantic-Detail Autoregressive Prediction}

\subsubsection{Naive AR Modeling of Semantic-Detail Codebook}
\yrexten{Our proposed Semantic-Detail Associated Dual-Codebook representation necessitates a new AR image generation manner tailored to learning the joint distribution over the two token sequences.}
A straightforward autoregressive approach to model the semantic-detail codebook treats each patch's semantic index $k_i$ and detail index $j_i$ as independent tokens in the sequence. Specifically, for an image with $m$ patches, we construct a token sequence $\{k_1, j_1, k_2, j_2, \ldots, k_m, j_m\}$, where $(k_i, j_i)$ encodes the semantic and detail representation for the $i$-th patch. The AR model then generates this doubled-length sequence, sequentially predicting each token conditioned on all previously generated tokens:
\begin{equation}
\begin{aligned}
    &p(\{k_1, j_1, \ldots, k_m, j_m\}) \\
    =& \prod_{i=1}^{m} p(k_i \mid \text{context}_{i})  \cdot p(j_i \mid k_i, \text{context}_{i}),
\end{aligned}
\end{equation}
\yrexten{where} $\text{context}_{i}$ denotes all tokens generated before patch $i$. While conceptually simple, this naïve approach incurs significant computational overhead due to the doubled sequence length, resulting in increased training and inference cost. Moreover, this method overlooks the hierarchical relationship between semantic and detail indices, as each is modeled at the same sequence level, which may hinder the model's ability to exploit conditional dependencies \yrexten{between detail and semantic tokens} within each patch.

\subsubsection{Semantic-Detail Autoregressive Prediction}

To address the drawbacks of the naive autoregressive modeling---namely, the doubling of sequence length and the neglect of the semantic-detail hierarchy---we propose a more efficient and effective approach for semantic-detail autoregressive prediction.

Instead of treating the semantic and detail indices as separate tokens, we introduce a \textbf{token fusion mechanism} that enables the joint modeling of both codebooks for each image patch without increasing the sequence length. \hut{Specifically, we extend the token embedding layer in the previous AR \yrexten{image generation} model~\cite{sun2024llamagen} into two distinct embedding layers: one for semantic tokens and one for detail tokens. For each patch $i$, we obtain its semantic embedding $Emb_s(k_i)$ and detail embedding $Emb_d(j_i)$ from their respective embedding layers, corresponding to the semantic and detail codebooks.} These embeddings are concatenated and subsequently projected into a unified patch representation $h_i$ by a multilayer perceptron (MLP):
\begin{equation}
    h_i = \text{MLP}\big([Emb_s(k_i);~Emb_d(j_i)]\big).
    \label{eq:token compression}
\end{equation}

The sequence of fused patch embeddings $\{h_1, \ldots, h_m\}$ is subsequently modeled by our AR backbone, \hut{outputting the contextualized hidden states $\{\hat{h}_1, \ldots, \hat{h}_m\}$, which efficiently model the context across the entire image.}

\noindent
\hut{\textbf{Hierarchical and Autoregressive Prediction with AR Head.}} The generative process for each spatial location requires predicting a structured pair of indices: one for \yrexten{the} semantic codebook and one for \yrexten{the} detail codebook. A straightforward strategy would involve employing two parallel prediction heads, such as MLPs, to independently map the transformer's output hidden state to logits for each codebook. However, this approach presumes \yrexten{the semantic and detail representations are independent.}
\yrexten{Consider generating an image patch containing an eye: the semantic concept ("an eye") fundamentally determines which high-frequency details are plausible, such as eyelash textures or iris patterns. Therefore,}
this \yrexten{independence} assumption is fundamentally misaligned with the inherent structure of visual data, where details are intrinsically conditioned on \yrexten{semantics.}

\hut{\yrexten{To address this issue, we perform} the prediction of semantic and detail tokens in a \textbf{hierarchical} and autoregressive manner, explicitly leveraging their natural dependence within each patch. To achieve this, we employ a dedicated \textbf{autoregressive (AR) head} (Fig.~\ref{fig:main framework of IAR2} \yrexten{(b)}), structured as a two-step process. First, the contextualized hidden state $\hat{h}_{i}$ output by the autoregressive backbone is used to predict the semantic token $k_{i+1}$ for the current patch. Specifically, $\hat{h}_{i}$ is projected to logits over the semantic codebook, yielding $p(k_{i+1} \mid h_{\leq i})=p(k_{i+1} \mid \hat{h}_i)$ where $h_{\leq i}$ denotes all embeddings for previous patches.}

\hut{Once the semantic token $k_{i+1}$ is predicted, we condition the prediction of the detail token $j_{i+1}$ on both the contextualized hidden state $\hat{h}_{i}$ and the newly predicted semantic token $k_{i+1}$. This is implemented by incorporating $k_{i+1}$ as an additional input token to the AR head, producing an enriched state to predict $j_{i+1}$ via $p(j_{i+1} \mid h_{\leq i}, k_{i+1})=p(j_{i+1} \mid \hat{h}_i,k_{i+1})$. This process can be formulated as:}
\begin{equation}
    p(k_{i+1}, j_{i+1} \mid h_{\leq i}) = p(k_{i+1} \mid h_{\leq i}) \cdot p(j_{i+1} \mid h_{\leq i}, k_{i+1}).
\end{equation}

\hut{By first establishing the semantic concept, the prediction of the detail token is conditioned on this strong prior, effectively narrowing the search space to only those details relevant to that concept. This two-stage conditional approach enables the model to capture the inherent semantic-to-detail hierarchy in visual data, where high-frequency details are generated in a manner consistent with the underlying semantics.
}

By adopting this hierarchical prediction strategy with token fusion, \hut{and explicitly modeling semantic–detail dependencies via the AR head, we reduce sequence length, preserve semantic-detail structure, and achieve significantly better training and inference efficiency compared to the naive AR baseline.}
Moreover, this approach aligns better with the compositional nature of patch representations in image modeling, enabling the model to fully leverage both high-level semantic \yrexten{information} and fine-grained details within a unified framework.

\subsubsection{Training Objective for Semantic-Detail Prediction}
The previous AR model is trained to predict the sequence of token indices using a cross-entropy loss. Reflecting our hierarchical prediction scheme, the total loss is a weighted sum of the losses for the semantic and detail tokens:
\begin{equation}
    \mathcal{L}_{\text{AR}} = \sum_{i=1}^{m} \left( -\lambda_{s}\log p(k_{i+1} | h_{\leq i}) -  \log p(j_{i+1} | h_{\leq i}, k_{i+1}) \right),
    \label{eq:loss_sd_ar_prediction}
\end{equation}
where $\lambda_{s}$ is a hyperparameter that balances the importance of correctly predicting the semantic \yrexten{information} versus fine-grained details. This formulation guides the model to first secure the correct semantic foundation before refining the details, effectively structuring the generative process.

\hutnew{By explicitly modeling the hierarchical dependency between semantic and detail tokens, this prediction scheme forms the architectural foundation of our generative head. Building upon this structural design, we further introduce a local-context enhancement mechanism in Sec.~\ref{sec:local_head} to explicitly capture spatial correlations within a local neighborhood, seamlessly integrating long-range global context with fine-grained local details during hierarchical prediction.}

\subsection{Local-Context Enhanced Autoregressive Head}
\label{sec:local_head}

\hutnew{Conventional autoregressive (AR) image generation models typically employ global attention within Transformer architectures to capture long-range, global dependencies across the entire image token sequence. While this type of full-sequence modeling is crucial for preserving holistic scene structure, it often overlooks the strong local correlations that are unique to visual data. Unlike natural language, where relationships between distant tokens are often essential for semantic understanding, the appearance of an image patch is primarily influenced by its immediate spatial neighbors. Effectively leveraging \textbf{local context} is therefore critical for enhancing texture continuity, boundary sharpness, and overall perceptual quality in image synthesis.
}

\hut{A naive solution is to inject local context modeling directly into the AR backbone module. However, this can introduce redundancy and may even interfere with the backbone’s capacity for global reasoning, thus diminishing its ability to model long-range structure. In our proposed framework, the AR head operates on a minimal input, typically just the backbone context and the semantic embedding, making limited use of the autoregressive modeling capacity. This underutilization further motivates a dedicated mechanism to \yrexten{exploit local context to enhance} local spatial coherence.}

\hutnew{To address these limitations, we propose the \textbf{Local-Context Enhanced Autoregressive Head}. Distinct from standard AR heads that predict tokens based solely on an isolated global hidden state, our module explicitly aggregates spatial information from the immediate neighborhood right before the semantic-detail classification. This targeted integration provides a powerful localized prior, ensuring that the prediction of the detail token is guided by both the global intent from the backbone and the concrete local patterns of adjacent patches. By clearly separating global macro-structure modeling in the backbone from local micro-detail enhancement in the AR head, our framework substantially improves generation fidelity, texture continuity, and edge sharpness without disrupting the long-range reasoning process.}

Specifically, for predicting the token at a given position \yrexten{(e.g., $i$-th token)}, we aggregate the hidden states from previously generated tokens \yrexten{within its} $k \times k$ local window \yrexten{(Fig.~\ref{fig:main framework of IAR2} (c))}. A naive concatenation of these local hidden states would be computationally \yrexten{expensive}. To maintain efficiency, we introduce a lightweight \textbf{context compression module}. Given a set of $N$ local contextualized hidden states, $\{\hat{h}_{\text{local}, n}\}_{n=1}^N$, the compression process is as follows:
\begin{enumerate}
    \item Each context vector $\hat{h}_{\text{local}, n}$ is passed through a shared \textit{compression network} (a small MLP) to reduce its dimensionality.
    \item The resulting low-dimensional vectors are concatenated along the feature dimension.
    \item A final \textit{FFN network} projects this concatenated vector back to the original hidden dimension, producing a single, fused \yrexten{local context} vector $\hat{h}_{ctx}$ that summarizes the local neighborhood.
\end{enumerate}
This process can be formulated as:
\begin{equation}
    \hat{h}_{ctx} = \text{FFN}(\text{Concat}(\{\text{Compress}(\hat{h}_{\text{local}, n})\}_{n=1}^{N})).
\end{equation}

The final prediction process integrates the global context from the Transformer with this compressed local context. 
\yrexten{As shown in Fig.~\ref{fig:main framework of IAR2} (b)}, the global context vector $\hat{h}_{i}$ and the local context vector $\hat{h}_{ctx}$ are fed into the \yrexten{AR} head. A lightweight attention mechanism is employed, where $\hat{h}_{i}$ acts as the query to attend to $\hat{h}_{ctx}$ (which serves as both key and value). 
\hut{This produces a refined, context-rich state $\tilde{h}_s=Head(\hat{h}_{ctx},\hat{h}_{i})$ that is used to predict the semantic token index $k_{i+1}={S\text{-}MLP(\tilde{h}_s)}$ by \yrexten{a} semantic MLP $S\text{-}MLP(\cdot)$.
Subsequently, the hidden state of this predicted semantic token is combined with the context-rich state to predict the detail token index $j_{i+1}={D\text{-}MLP(\tilde{h}_d)}$, where $\tilde{h}_d=Head(\tilde{h}_s,\hat{h}_{ctx},\hat{h}_{i})$ and $D\text{-}MLP(\cdot)$ is \yrexten{a} detail MLP.} 
This integrated design ensures that predictions are guided by both long-range dependencies captured in $\hat{h}_{i}$ and the fine-grained local structure summarized in $\hat{h}_{ctx}$, while strictly adhering to the desired semantic-to-detail generative hierarchy.

\begin{figure}[t]
\centering
\includegraphics[width=0.48\textwidth]{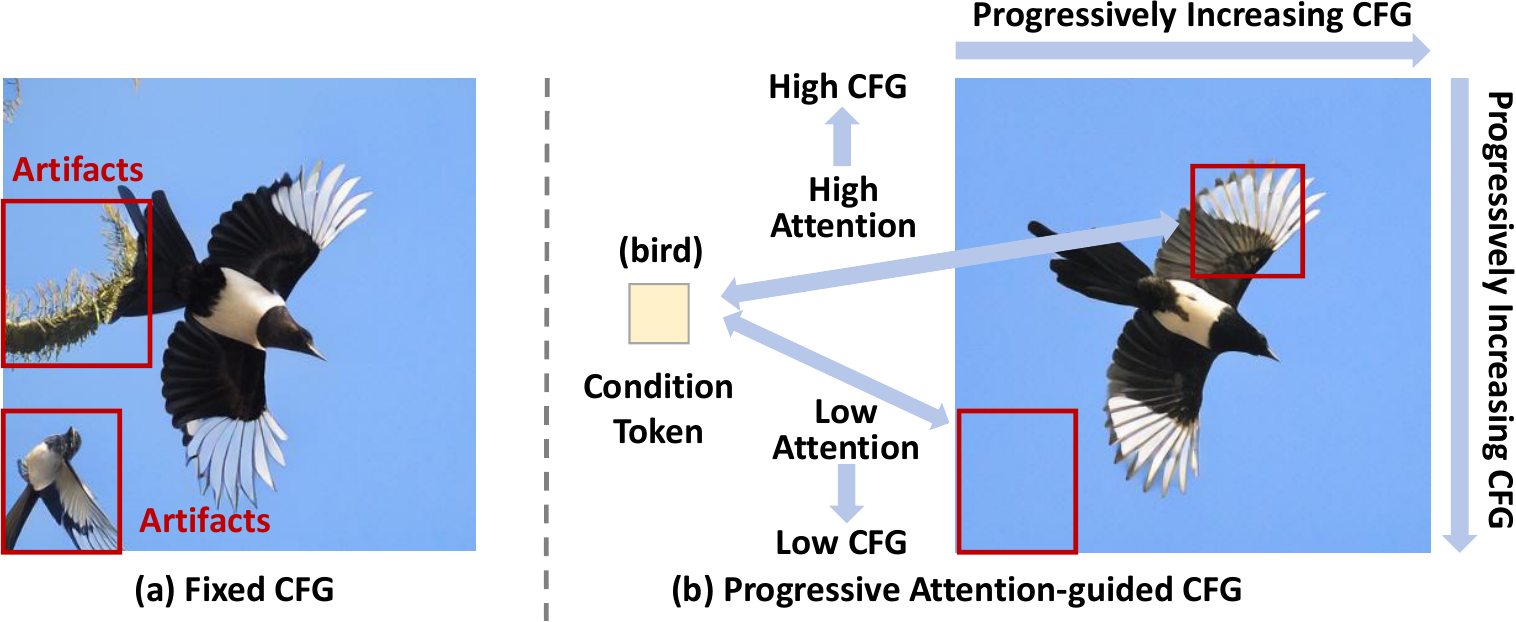}
\caption{Comparison between conventional fixed CFG (a) and our progressive attention-guided CFG (b). The conventional approach applies a uniform guidance scale, leading to artifacts in background regions. In contrast, our method adaptively modulates guidance based on \yrexten{spatial relevance (}attention scores) \yrexten{and sequential progress}, reducing the scale for the background to prevent artifacts while strengthening it for the subject to enhance semantic fidelity, 
\yrexten{and intensifying the strength as generation proceeds.}}
\label{fig:attention-guided-cfg}
\end{figure}

\subsection{Progressive Attention-Guided CFG}

Classifier-Free Guidance (CFG) is a foundational technique for enhancing conditional alignment and perceptual quality in autoregressive image generation models. \hut{It operates by amplifying the conditioning signal, such as text prompts or class types, thus steering the generation process toward the desired attributes. In this work, the conditioning signal specifically refers to the class type.}
However, the conventional approach of applying a single, fixed CFG scale globally is suboptimal because it overlooks two critical dynamics: 
\textbf{1) Spatial Uniformity:} the relevance of the conditioning signal is not uniform across \yrexten{different regions of} an image, and a strong guidance scale beneficial for the main subject can introduce artifacts in background regions that are semantically \yrexten{less related} to the condition. 
\textbf{2) Sequential Staticity:} 
\hut{in the autoregressive process, the influence of the external condition is not constant. As \yrexten{more patches of the image are generated,} the accumulated internal context grows stronger, potentially overshadowing the external condition. A fixed guidance scale fails to counteract this dynamic shift, often proving too weak in later stages \yrexten{of the generation process}, which \yrexten{can} cause semantic misalignment.}

To address these limitations, we propose \textbf{Progressive Attention-Guided CFG (PAG-CFG)}, a novel mechanism that modulates the \yrexten{CFG} guidance strength dynamically. Our method first uses \textbf{attention guidance} to tailor the CFG scale to the spatial content of the image, and then introduces a \textbf{progressive schedule} to adapt it to the sequential stage of generation. This ensures that the guidance is applied precisely where and when it is most needed, significantly improving both conditional \yrexten{alignment} and overall image quality.

\noindent\textbf{Classifier-Free Guidance Preliminaries.}
Standard CFG adjusts the model's output logits by blending the conditional and unconditional predictions. For the $i$-th token, the guided logits $l_{cfg}$ are computed as:
\begin{equation}
    l_{cfg}(y_i | y_{<i}, c) = l_u(y_i | y_{<i}) + s \cdot (l_c(y_i | y_{<i}, c) - l_u(y_i | y_{<i})),
    \label{eq:standard_cfg}
\end{equation}
where $s$ is the static, global guidance scale, $c$ is the condition, and $l_c$ and $l_u$ are the conditional and unconditional logits, respectively. Our goal is to replace the fixed scale $s$ with a dynamic, per-token scale $s_i$.

\noindent\textbf{Attention-Guided Spatial Modulation.}
To solve the problem of \textit{Spatial Uniformity}, we make the guidance strength proportional to the semantic relevance between the generated token and the condition.
\yrexten{This adaptive guidance strategy ensures that the conditional influence is concentrated on semantically relevant regions. For tokens strongly related to the condition (like those in the foreground region), we apply a higher CFG scale to powerfully steer the generation. Conversely, for tokens in irrelevant areas (e.g., the background), the guidance is weakened to avoid introducing unnecessary constraints.}

The attention mechanism within the Transformer is perfectly suited for this task, as its scores inherently quantify this relationship. \yrexten{Therefore,} for each token $y_i$, we aggregate its attention weights $\mathbf{A}_i \in \mathbb{R}^{L_c}$ towards the $L_c$ condition tokens to derive a single relevance score $\alpha_i = \text{Aggregate}(\mathbf{A}_i) \in [0, 1]$. This allows CFG to act as a "semantic spotlight," intensifying guidance on tokens corresponding to the main subject while applying a lighter touch to the background, thereby mitigating artifacts.

\noindent\textbf{Progressive Sequential Scheduling.}
The "progressive" aspect of our method addresses \textit{Sequential Staticity} by introducing a scheduling mechanism that adapts the guidance strength \yrexten{as} the generative \yrexten{progresses}. The core insight is that the \yrexten{strength} of CFG should evolve \yrexten{throughout} the generation process \yrexten{of $M$ tokens}: 
\hut{as more tokens are generated, the accumulated internal context (i.e., the preceding tokens) becomes increasingly influential, potentially overshadowing the external condition. To counteract this drift and maintain strong conditional alignment, a progressively stronger guidance signal is required in the later stages.}
Consequently, we employ a schedule that gradually \textit{increases} the base guidance scale from a starting value $s_{start}$ to a final value $s_{end}$ over the course of generating $M$ tokens. The scheduled base scale for token $i$ is:
\begin{equation}
    s'_{i} = s_{start} + (s_{end} - s_{start}) \cdot \frac{i}{M}.
\end{equation}
This strategy makes the CFG process temporally aware, shifting its focus from coarse composition to fine-grained, condition-aligned refinement.

\noindent\textbf{Final CFG Formulation.}
By combining the attention-guided spatial factor $\alpha_i$ with the progressive sequential schedule $s'_{i}$, we derive a final adaptive scale $s_i$ that is \yrexten{aware of} both "what" is being generated (spatial relevance) and "when" it is being generated (sequential progress):
\begin{equation}
    s_i = s'_{i} \cdot \alpha_i = \left( s_{start} + (s_{end} - s_{start}) \cdot \frac{i}{M} \right) \cdot \alpha_i.
\end{equation}
Substituting this dynamic scale $s_i$ for $s$ in Equation~(\ref{eq:standard_cfg}) yields our final PAG-CFG. 
\hutnew{To enable this inference-time capability, the model is trained by randomly replacing the conditioning signal with a null token with a 10\% probability, thereby allowing the network to jointly model both conditional and unconditional distributions. During inference, this dual-modulated approach empowers the CFG to effectively and adaptively adjust its weight according to both the attention map and the current generation progress, thereby achieving fine-grained and context-aware control over the conditioning strength. As a result, the method significantly improves conditional alignment for the subject while preserving the natural quality of the entire image.}

\section{Experiments}
\label{sec:experiments}

\subsection{Experiment Settings}

\textbf{Implementation Details.}
Our autoregressive model adopts the LlamaGen~\cite{sun2024llamagen} \yrexten{as the base model}, which consists of a stack of Transformer layers. To encode the spatial location of image patches, we employ 2D Rotary Position Embeddings (2D-RoPE). We conduct experiments across a range of model scales, from 100M to 1.5B parameters, to assess the scalability of our proposed methods. Our custom autoregressive head is implemented as a lightweight five-layer Transformer, with its hidden dimension and number of attention heads configured to match those of the base model.
All models are trained and evaluated on the ImageNet dataset~\cite{deng2009imagenet}. To ensure a fair and direct comparison, we strictly adhere to the training protocol established by LlamaGen. This includes using the identical batch size, the AdamW optimizer with its corresponding hyperparameters ($\beta_1, \beta_2, \epsilon$), and training all models for a total of 300 epochs. 
\yrexten{More detailed hyperparameter settings are provided in the Supplementary Material.}

\noindent\textbf{Evaluation Metrics.}
To comprehensively evaluate the generative performance of our models, we synthesize a total of 50,000 images for each model, sampling across all 1,000 classes from the ImageNet validation set. We then compute the following standard metrics:
\begin{itemize}
    \item \textbf{Fr\'{e}chet Inception Distance (FID)}~\cite{fid} measures the similarity between the distributions of real and generated images in the feature space of an InceptionV3 network. A lower FID score indicates higher visual quality and better fidelity to the training data distribution.
    \item \textbf{Inception Score (IS)}~\cite{inception_score} evaluates both the quality (clarity) and diversity of generated images. A higher IS suggests that the model generates more distinct and recognizable objects.
    \item \textbf{Precision and Recall}~\cite{precision_and_recall} are used to assess class-conditional generation. Precision measures the fidelity of generated samples (what fraction are realistic), while Recall measures diversity (what fraction of the real data distribution is covered). Higher values for both are desirable.
\end{itemize}

\begin{table}[t]
\caption{Ablation study on different codebook sizes. CB1 and CB2 denote the codebook sizes of the semantic and detail codebooks, respectively.}
\label{tab:codebook_ablation1}
\centering
\resizebox{0.47\textwidth}{!}{
\begin{tabular}{cc|ccc|cccc}
\toprule
\multicolumn{2}{c|}{Codebook Size} & \multicolumn{3}{c|}{Reconstruction} & \multicolumn{4}{c}{Generation} \\
\multicolumn{1}{c}{CB1} & \multicolumn{1}{c|}{CB2} & rFID$\downarrow$ & PSNR$\uparrow$ & SSIM$\uparrow$ & gFID$\downarrow$ & IS$\uparrow$ & Precision$\uparrow$ & Recall$\uparrow$ \\
\midrule
128  & 4096  & 1.73 & 20.91 & 0.67  &6.06 & 188.9 & \textbf{0.84} & \textbf{0.42}  \\
\textbf{256}  & \textbf{4096}  & 1.72 & \textbf{20.95}  & 0.67  &  \textbf{6.05} & \textbf{208.0} & \textbf{0.84} & 0.40 \\
512  & 4096  & \textbf{1.67} & 20.23  & \textbf{0.68}  & 6.14  & 185.7 & 0.83 &0.41  \\
\midrule
256  & 2048  & 2.02 & 20.93 &  \textbf{0.67} & 6.15 & 197.4 & \textbf{0.85} &\textbf{0.40}  \\
\textbf{256}  & \textbf{4096}  & 1.72 & 20.95  & \textbf{0.67}   &\textbf{6.05} & \textbf{208.0} & 0.84 & \textbf{0.40} \\
256  & 8192  & \textbf{1.69} & \textbf{21.00}  & \textbf{0.67}  & 6.18 & 191.1 & \textbf{0.85} & \textbf{0.40} \\
\bottomrule
\end{tabular}}
\end{table}

\subsection{Impact of Codebook Configurations}

The capacity and organization of the semantic and detail codebooks in our joint quantization framework play a central role in balancing reconstruction fidelity and generative expressiveness. To systematically investigate their effect, we conducted a series of experiments varying the size of each codebook individually while keeping the others fixed. Table~\ref{tab:codebook_ablation1} summarizes both reconstruction (rFID, PSNR, SSIM) and generation metrics (gFID, IS, Precision, Recall) under different configurations.

We observe that increasing the size of the semantic codebook from 128 to 256 entries consistently improves \yrexten{reconstruction} \hut{PSNR, reflecting better preservation of global semantic information, while also achieving a better generation quality (better gFID and IS)}. However, further enlarging to 512 
\yrexten{degrades generation quality compared to a smaller semantic codebook size, and brings only marginal reconstruction gains. This}
\hut{
indicates that a \yrexten{semantic} codebook of 256 entries provides sufficient semantic capacity, with further increases offering no meaningful improvement.}

For the detail codebook, increasing capacity from 2048 to 4096 entries markedly enhances both reconstruction fidelity and generative quality, as indicated by higher PSNR, IS scores, and balanced Precision/Recall. Nevertheless, \yrexten{further increasing} the detail codebook size to 8192 offers only modest additional improvements, which even results in a decrease in gFID.

Overall, the combination of a moderately sized semantic codebook (256 entries) and a sufficiently large detail codebook (4096 entries) delivers the most favorable trade-off: low rFID, strong PSNR and SSIM for reconstruction, and robust generation quality. 
\hut{Further increasing the codebook sizes provides only marginal gains in reconstruction at the cost of degraded generation performance, underscoring that our default configuration strikes an effective balance between model capacity and overall efficacy.}

\begin{table}[t]
\caption{Comparison between different types of image generation model on class-conditional ImageNet 256$\times$256 benchmark with FID, IS, precision, and recall. $^*$ indicates reject sampling.
}
\label{tab:main}
\centering
\resizebox{0.48\textwidth}{!}{
\begin{tabular}{c|lc|cccc}
\toprule
Type & Model & \#Para. & FID$\downarrow$ & IS$\uparrow$ & Precision$\uparrow$ & Recall$\uparrow$  \\
\midrule
\multirow{3}{*}{GAN}   & BigGAN~\cite{biggan}  & 112M   & 6.95  & 224.5       & 0.89 & 0.38 \\
 & GigaGAN~\cite{gigagan}  & 569M    & 3.45  & 225.5       & 0.84 & 0.61  \\
 & StyleGan-XL~\cite{stylegan-xl} & 166M    & 2.30  & 265.1       & 0.78 & 0.53   \\
\midrule
\multirow{4}{*}{Diffusion} & ADM~\cite{adm}  & 554M       & 10.94 & 101.0        & 0.69 & 0.63    \\
 & CDM~\cite{cdm}   & $-$       & 4.88  & 158.7       & $-$  & $-$   \\
 & LDM-4~\cite{ldm} & 400M     & 3.60  & 247.7       & $-$  & $-$  \\
 & DiT-XL/2~\cite{dit}  & 675M  & 2.27  & 278.2       & 0.83 & 0.57   \\
\midrule
\multirow{2}{*}{Mask.} & MaskGIT~\cite{maskgit}  & 227M   & 6.18  & 182.1        & 0.80 & 0.51  \\
 & MaskGIT$^*$~\cite{maskgit} & 227M\    & 4.02  & 355.6        & $-$ & $-$ \\
\midrule
\multirow{1}{*}{VAR.} &VAR-d30~\cite{VAR}  & 2.0B &1.92  &323.1 & 0.82 & 0.59 \\
\midrule
\multirow{16}{*}{AR} 
& VQGAN~\cite{vqgan} & 227M & 18.65 & 80.4         & 0.78 & 0.26    \\
 & VQGAN~\cite{vqgan}    & 1.4B   & 15.78 & 74.3   & $-$  & $-$     \\
 & VQGAN$^*$~\cite{vqgan}  & 1.4B  & 5.20  & 280.3  & $-$  & $-$     \\
 & ViT-VQGAN~\cite{vit-vqgan} & 1.7B & 4.17  & 175.1  & $-$  & $-$        \\
 & ViT-VQGAN$^*$~\cite{vit-vqgan}& 1.7B  & 3.04  & 227.4  & $-$  & $-$     \\
 & RQTran.~\cite{rq}       & 3.8B  & 7.55  & 134.0  & $-$  & $-$     \\
 & RQTran.$^*$~\cite{rq}    & 3.8B & 3.80  & 323.7  & $-$  & $-$    \\
&\hutnew{UniTok}~\cite{ma2026unitok} & 1.4B & 2.51 & 216.7 & 0.82 & 0.57 \\
&\hutnew{NPP-XXL}~\cite{npp} & 1.4B & 2.54 & 286.1 & 0.84 & 0.56 \\
&\hutnew{PAR-XXL}~\cite{wang2025parallelized} & 1.4B & 2.35 & 263.2 & 0.82 & 0.57 \\
& LlamaGen-B~\cite{sun2024llamagen}& 111M & 5.46 & 193.6 & 0.83 & 0.45\\
 & LlamaGen-L~\cite{sun2024llamagen}   & 343M & 3.29 & 227.8 & 0.82 & 0.53 \\
 & LlamaGen-XL~\cite{sun2024llamagen}  & 775M & 2.63 & 244.1 & 0.81 & 0.58 \\
 & LlamaGen-XXL~\cite{sun2024llamagen}  & 1.4B & 2.34 & 253.9 & 0.80 & 0.59 \\
& IAR-B~\cite{hu2025iar}& 111M & 5.14 & 202.0 & 0.85 & 0.45\\
 & IAR-L~\cite{hu2025iar}  & 343M & 3.18 & 234.8 & 0.82 & 0.53 \\
 & IAR-XL~\cite{hu2025iar}  & 775M & 2.52 & 248.1 & 0.82 & 0.58 \\
& IAR-XXL~\cite{hu2025iar}   & 1.4B & 2.19 & 265.6 & 0.81 & 0.58 \\
\midrule
\multirow{5}{*}{AR} 
& IAR2-B& 143M &  4.06 & 219.6 & 0.84 & 0.47\\
 & IAR2-L  & 408M &2.57 & 276.2 & 0.83 & 0.55 \\
 & IAR2-XL  & 884M &2.10 & 286.4 & 0.80 & 0.59   \\
& IAR2-XXL  & 1.5B &1.76 & 279.5&0.80 &0.62 \\
& IAR2-XXL$^*$  & 1.5B &\textbf{1.50} & 282.7&0.80 &0.63 \\
& IAR2-XXL$^*$  & 1.5B &2.10 & \textbf{360.6}&0.78 &0.61 \\

\bottomrule
\end{tabular}}
\vspace{-0.15in}
\end{table}

\begin{table*}[t]
\caption{Comparison with LlamaGen and IAR across different image tokens and model sizes. Following LlamaGen, we only train XL and XXL versions on $16\times 16$ tokens for 50 epochs, while all other models are trained for 300 epochs. For each metric, the best result (within the same token size \& epoch) is highlighted in \textbf{bold}, and the second best is \underline{underlined}.}
\label{tab:comparison with baselines}
\centering
\resizebox{0.8\textwidth}{!}{
\begin{tabular}{c|lc|cccc|cccc}
\toprule
\multirow{2}{*}{Tokens} & \multicolumn{1}{c}{\multirow{2}{*}{Model}} & \multicolumn{1}{c}{\multirow{2}{*}{\#Para.}} & \multicolumn{4}{c}{50 epoch} & \multicolumn{4}{c}{300 epoch} \\
& \multicolumn{1}{c}{} & \multicolumn{1}{c}{} & \multicolumn{1}{c}{FID$\downarrow$} & \multicolumn{1}{c}{IS$\uparrow$} & \multicolumn{1}{c}{Precision$\uparrow$} & \multicolumn{1}{c}{Recall$\uparrow$} & \multicolumn{1}{c}{FID$\downarrow$} & \multicolumn{1}{c}{IS$\uparrow$} & \multicolumn{1}{c}{Precision$\uparrow$} & \multicolumn{1}{c}{Recall$\uparrow$} \\
\midrule

\multirow{12}{*}{16 $\times$ 16} 
& LlamaGen-B & 111M & 7.22 & \underline{178.3} & \textbf{0.86} & 0.38 & 5.46 & 193.6 & \underline{0.84} & \underline{0.46} \\
& LlamaGen-L & 343M & 4.20 & \underline{200.0} & \textbf{0.82} & \underline{0.51} & 3.80 & 248.3 & \underline{0.83} & \underline{0.52} \\
& LlamaGen-XL & 775M & 3.39 & 227.1 & 0.81 & 0.54 & - & - & - & - \\
& LlamaGen-XXL & 1.4B & 3.09 & 253.6 & 0.83 & 0.53 & - & - & - & - \\
\cline{2-11}
& IAR-B & 111M & \underline{6.90} & \textbf{179.2} & \textbf{0.86} & \underline{0.40} & \underline{5.14} & \underline{202.0} & \textbf{0.85} & 0.45 \\
& IAR-L & 343M & \underline{4.10} & \textbf{207.1} & \textbf{0.82} & \underline{0.51} & \underline{3.40} & \underline{271.3} & \textbf{0.84} & 0.51 \\
& IAR-XL & 775M & \underline{3.36} & \underline{228.9} & \underline{0.82} & 0.54 & - & - & - & - \\
& IAR-XXL & 1.4B & \underline{3.01} & \underline{257.4} & \underline{0.83} & 0.53 & - & - & - & - \\
\cline{2-11}
& IAR2-B & 143M & \textbf{5.61} & 177.0 & 0.84 & \textbf{0.43} & \textbf{4.06} & \textbf{219.6} & \underline{0.84} & \textbf{0.47}  \\
& IAR2-L & 408M & \textbf{3.77} & 192.6 & 0.79 & \textbf{0.54} & \textbf{2.57} & \textbf{276.2} & \underline{0.83} & \textbf{0.55} \\
& IAR2-XL & 884M & \textbf{2.64} & \textbf{241.8} & 0.81 & \underline{0.56} & - & - & - & - \\
& IAR2-XXL & 1.5B & \textbf{2.30} & \textbf{263.3} & 0.81 & \textbf{0.58} & - & - & - & - \\
\midrule

\multirow{12}{*}{24 $\times$ 24} 
& LlamaGen-B & 111M & 8.31 & \underline{154.7} & \underline{0.84} & 0.38 & 6.09 & 182.5 & \underline{0.84} & \underline{0.42} \\
& LlamaGen-L & 343M & 4.61 & 191.4 & \underline{0.82} & 0.50 & 3.29 & 227.8 & \textbf{0.82} & \underline{0.53} \\
& LlamaGen-XL & 775M & 3.24 & \underline{245.7} & \textbf{0.83} & 0.53 & 2.63 & 244.1 & \underline{0.81} & \underline{0.58} \\
& LlamaGen-XXL & 1.4B & \underline{2.89} & 236.2 & 0.80 & \textbf{0.56} & {2.34} & 253.9 & \textbf{0.81} & \underline{0.60} \\
\cline{2-11}
& IAR-B & 111M & \underline{7.80} & {153.3} & \underline{0.84} & \textbf{0.39} & \underline{5.77} & \underline{192.5} & \textbf{0.85} & \underline{0.42} \\
& IAR-L & 343M & \underline{4.35} & \underline{197.2} & 0.81 & \underline{0.51} & \underline{3.18} & \underline{234.8} & \textbf{0.82} & \underline{0.53} \\
& IAR-XL & 775M & \underline{3.15} & 228.8 & \underline{0.81} & \underline{0.54} & \underline{2.52} & \underline{248.1} & \textbf{0.82} & \underline{0.58} \\
& IAR-XXL & 1.4B & \underline{2.87} & \underline{249.9} & \textbf{0.82} & \textbf{0.56} & \underline{2.19} & \underline{265.6} & \textbf{0.81} & 0.58 \\
\cline{2-11}
& IAR2-B & 143M & \textbf{6.90} & \textbf{174.8} & \textbf{0.85} & \textbf{0.39}  & \textbf{4.80} & \textbf{211.8} & \underline{0.84} & \textbf{0.45} \\
& IAR2-L & 408M & \textbf{4.05} & \textbf{236.1} & \textbf{0.84} & \textbf{0.48} & \textbf{2.76} & \textbf{257.9} & \underline{0.81} & \textbf{0.56} \\
& IAR2-XL & 884M & \textbf{2.77} & \textbf{251.1} & 0.80 & \textbf{0.56} & \textbf{2.10} & \textbf{286.4} & 0.80 & \textbf{0.59} \\
& IAR2-XXL & 1.5B & \textbf{2.74} & \textbf{279.8} & \textbf{0.82} & \textbf{0.56} & \textbf{1.76} & \textbf{279.5} & 0.80 & \textbf{0.62}  \\
\bottomrule
\end{tabular}}
\end{table*}

\subsection{Comparison Results on Image Generation}

\textbf{Comparison with the State-of-the-arts.}
We conduct a comprehensive comparison of our IAR2 model against representative approaches across four major paradigms: GAN-based methods~\cite{biggan,gigagan,stylegan-xl}, diffusion-based methods~\cite{adm,cdm,ldm,dit}, mask-prediction methods~\cite{maskgit}, and autoregressive methods~\cite{vqgan,vit-vqgan,rq,sun2024llamagen,VAR,hu2025iar,ma2026unitok,npp,wang2025parallelized} on the class-conditional ImageNet benchmark~\cite{deng2009imagenet}. As summarized in Table~\ref{tab:comparison with baselines} \hutnew{($^*$ denotes reject sampling)}, IAR2 achieves the state-of-the-art performance, reaching an FID of 1.50 and an IS of 286.4, surpassing all the existing baselines. 

Several observations can be made. First, while GANs and diffusion models have historically dominated ImageNet generation, our IAR2 consistently delivers superior fidelity and diversity. Notably, compared to DiT-XL/2~\cite{dit}, the strongest diffusion baseline, IAR2 \yrexten{improves} FID from 2.27 to 1.50 and improves IS from 278.2 to 286.4, highlighting the scalability of autoregressive transformers when equipped with an effective design. 
Second, relative to recent autoregressive methods such as LlamaGen~\cite{sun2024llamagen} and IAR~\cite{hu2025iar}, IAR2 achieves \yrexten{steady gains} across all model sizes, demonstrating the robustness of our improvements in both semantic modeling and token-level generation. 
\hut{Thirdly, our work marks a significant leap in both generative performance and computational accessibility. IAR2-XXL sets a new state-of-the-art FID of \textbf{1.50} with a 1.5B parameter model, surpassing the larger 2.0B VAR model~\cite{VAR}. More strikingly, this was accomplished on a remarkably modest hardware setup of 32 GPUs, in contrast to the 256-GPU cluster used to train VAR. This demonstrates that our framework is not only more parameter-efficient but also substantially more resource-efficient, making state-of-the-art image generation more attainable.}

\noindent\textbf{More Comparison with LlamaGen and IAR.} 
We conduct more quantitative comparisons with both LlamaGen and IAR on different model sizes, training epochs, and image token numbers. The results in \yrexten{Table}~\ref{tab:comparison with baselines} highlight several noteworthy trends. 
First, across all tested model scales, IAR2 delivers consistent improvements over LlamaGen, with FID reduced by as much as 0.8--1.2 and IS increased by 20–40. 
This gap remains stable from small models to billion-parameter variants, suggesting that the architectural changes in IAR2 provide benefits beyond what can be achieved by simply scaling up model size. 
Compared with IAR, which already demonstrated clear advantages over LlamaGen, IAR2 pushes the performance further: the \yrexten{new method designs} introduced here address not only reconstruction fidelity but also diversity, resulting in both lower FID and higher IS. 
An additional observation is that the gains hold even at the more challenging $24\times24$ tokenization, corresponding to $384\times384$ resolution, where error accumulation typically hampers autoregressive methods. 
The fact that IAR2 \yrexten{maintains its superiority} under this setting indicates that it generalizes more robustly across resolutions. 
We also find that the relative advantages of IAR2 persist under both short (50 epochs) and long (300 epochs) training schedules, showing that the improvements are not merely a byproduct of extended optimization but rather stem from the underlying design. 

Taken together, these comparisons demonstrate that IAR2 represents a meaningful step forward from both LlamaGen and IAR. 
Although the three models share a similar autoregressive philosophy, IAR2 introduces methodological differences that \yrexten{lead to} measurable improvements in fidelity, diversity, and scalability, making it a stronger foundation for future work in LLM-based visual generation.

\subsection{\yrexten{Ablation Study on Core Components}}

\begin{table}[t]
\caption{Ablation study evaluating the effectiveness of \yrexten{core components:} Semantic–Detail Decoupling, Local-Context Enhancement, and PAG-CFG. The best result is highlighted in \textbf{bold}, and the second best is \underline{underlined}.}
\label{tab:ablation_study_full_modules}
\centering
\renewcommand{\arraystretch}{1.2}
\resizebox{0.47\textwidth}{!}{ 
\begin{tabular}{ccc|cccc}
\toprule
 \makecell[c]{Semantics-Detail\\ Association} & \makecell[c]{Local-Context\\ Enhancement} & \makecell[c]{PAG-CFG} &  FID$\downarrow$ & IS$\uparrow$ & Precision$\uparrow$ & Recall$\uparrow$  \\
\midrule
 &  &  & 6.74 & 169.2 & 0.82 & \underline{0.41}\\
\checkmark&  &  & 6.29 & 186.3 & 0.84 & \underline{0.41}  \\
  & \checkmark &  &  6.57 & 175.6 & 0.83 & 0.40 \\

 \checkmark& \checkmark &  &  6.18 &187.9  &   \textbf{0.85} & 0.40  \\
 \checkmark&  & \checkmark & \underline{6.04} & \textbf{196.9} & \underline{0.84} & \textbf{0.42}  \\
 \checkmark & \checkmark & \checkmark & \textbf{5.89} & \underline{192.8}  & \textbf{0.85} & 0.40  \\
\bottomrule
\end{tabular}}
\end{table}

We conduct a comprehensive ablation study to dissect the individual contributions of our core components: (1) Semantic-Detail Associated Dual Codebook; (2) Local Context enhancement, and (3) Progressive Attention-Guided CFG (PAG-CFG). 
\yrexten{The experiments are conducted for 100 epochs under 143M parameters (B version).}
\yrexten{The} results \yrexten{are} summarized in Table~\ref{tab:ablation_study_full_modules}. 
\hut{Our analysis begins with a baseline model \yrexten{that removes all three core components.} 
This model yields a high FID of 6.74, establishing a clear lower bound on performance. 
The introduction of our first core component, the \textbf{Semantic-Detail Associated Dual Codebook}, provides a dramatic improvement, reducing the FID to 6.29. This substantial gain underscores the critical role of our dual codebook in capturing both high-level semantics and fine-grained details, establishing a new, strong baseline upon which we evaluate the remaining modules. 
From this strong baseline}, integrating the Local-Context Enhanced Autoregressive Head further improves the FID to 6.18 by enhancing local coherence. 
More notably, the Progressive Attention-Guided CFG (PAG-CFG) yields a significant single-component gain when added to the baseline \yrexten{with semantic-detail codebook}, reducing the FID to 6.04 and boosting the IS to a remarkable 196.9, underscoring its effectiveness in strengthening semantic alignment and sample diversity. 
Finally, the full model, which integrates all three components, demonstrates their synergistic effect by achieving the best overall performance. It obtains the lowest FID of 5.89 and the highest precision of 0.85. Although its IS of 192.8 is slightly surpassed by the PAG-CFG variant, the superior FID score confirms a significant gain in image fidelity and realism. These results empirically validate that our proposed \yrexten{core} components are complementary and collectively lead to a state-of-the-art synthesis capability.

\subsection{Analysis on the Generation Hyperparameters}

\begin{figure*}[t]
\centering
\includegraphics[width=0.98\textwidth]{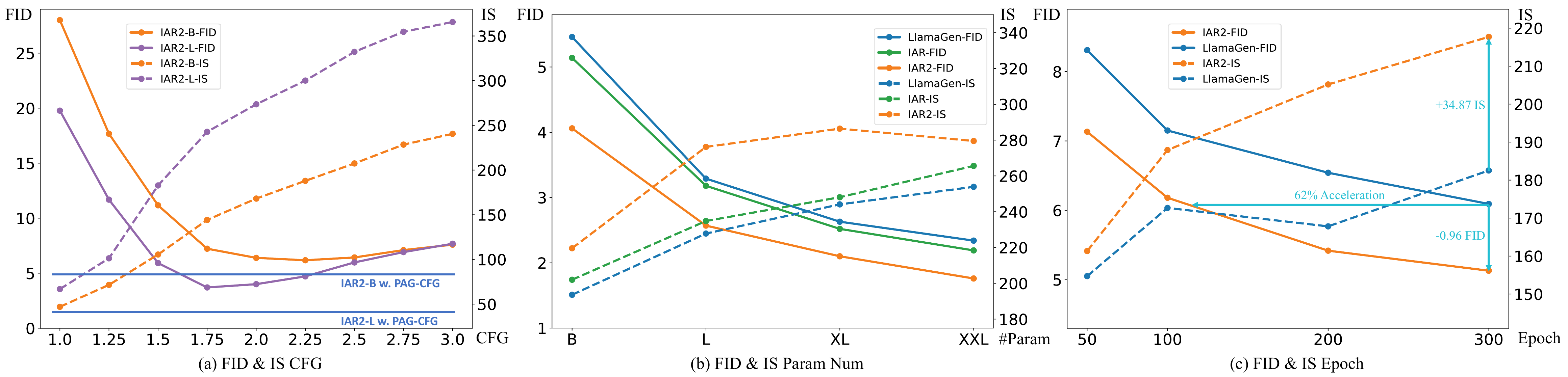}
\caption{Analysis on the Generation Hyperparameters: (a) CFG strength; (b) Parameter number; and (3) Training epoch. }
\label{fig:more comparison}
\end{figure*}

\noindent\textbf{Effectiveness of Progressive Attention-Guided CFG.}
\hut{To evaluate the effectiveness of our Progressive Attention-Guided CFG (PAG-CFG), we compare its performance against the conventional static CFG method across a range of guidance scales. The results for both IAR2-B and IAR2-L models are presented in Fig.~\ref{fig:more comparison} (a).
Our analysis of the static CFG reveals a clear performance trend: As the guidance scale increases from 1.0, both FID and IS scores improve, indicating enhanced sample quality. This improvement peaks within an optimal range of approximately 1.75 to 2.0, where the static approach achieves its best possible balance between fidelity and class-conditional alignment. Beyond this point, further increasing the guidance scale leads to 
\yrexten{worse FID scores,}
as overly strong, uniform guidance begins to degrade sample quality.}

\hut{Crucially, when compared with our PAG-CFG (the horizontal lines in Fig.~\ref{fig:more comparison} (a)), it demonstrates a substantial leap in performance. For both IAR2-B and IAR2-L models, our PAG-CFG significantly outperforms the best static CFG. This significant improvement underscores the advantage of our dynamic guidance strategy. By adaptively modulating guidance strength based on semantic context rather than applying a fixed scale globally, PAG-CFG achieves a superior synthesis quality that is unattainable with the conventional static \yrexten{CFG} approach. This experiment empirically validates that PAG-CFG is a more powerful and effective mechanism for guiding high-fidelity autoregressive image generation.}

\noindent\textbf{Effect of Model Size on Generation Quality.}
Fig.~\ref{fig:more comparison}(b) shows the relationship between model parameter size and generation performance, as measured by FID and IS, for LlamaGen, IAR, and our proposed IAR2 across four parameter scales (B, L, XL, XXL).
Across all models, increasing \yrexten{the} parameter count consistently reduces FID, indicating improved image fidelity with larger model capacity. This trend is most pronounced for IAR2, which achieves the lowest FID values at each scale. \hut{Notably, IAR2 consistently outperforms both IAR and LlamaGen by a substantial margin across all parameter scales.}
For instance, at the XXL scale, IAR2 reaches an FID \yrexten{of 1.76}, substantially lower than the corresponding values for the other two methods.
\hut{Similarly, IS scores improve with model size across all frameworks, \yrexten{with} our IAR2 consistently achieves the highest scores.
\yrexten{While the performance saturates at the largest scale, our IAR2 still maintains a leading advantage over}
both IAR and LlamaGen.} 
This demonstrates its superior ability to generate diverse and high-quality samples as model capacity grows. IAR2 consistently achieves the highest IS across scales, with gains most evident at larger model sizes.
Overall, these results highlight the strong scalability of IAR2: it not only benefits from increasing parameters but also \yrexten{demonstrates} superior performance compared to previous approaches at every scale. The pronounced improvements in both FID and IS indicate that IAR2 leverages its architectural innovations to maximize generative quality as model \yrexten{size} grows, setting state-of-the-art performance across standard parameter configurations.

\noindent\textbf{Effect of Training Epochs on Generation Quality.}
Fig.~\ref{fig:more comparison}(c) illustrates the evolution of FID and IS metrics for IAR2 and LlamaGen as the number of training epochs increases from 50 to 300. \hut{For a fair comparison, all the results are sampled with a fixed CFG=2.25.}
\hut{Throughout the entire training \yrexten{process}, IAR2 demonstrates a remarkably consistent and stable improvement. Its FID score steadily decreases while its IS score monotonically increases up to 300 epochs, indicating a sustained enhancement in both image fidelity and diversity. In comparison, while LlamaGen also achieves a progressively lower FID, its IS score exhibits some \yrexten{fluctuation} during the intermediate stages of training.}
Notably, at 300 epochs, IAR2 achieves a large margin of improvement over LlamaGen, with a reduction of 0.96 in FID and an increase of 34.87 in IS. This demonstrates not only the superior final generation quality of IAR2 but also its greater training efficiency. As highlighted in the figure, IAR2 reaches a strong FID at a much earlier epoch, achieving a 62\% acceleration in convergence compared to LlamaGen.
\hut{Overall, these findings confirm that IAR2 is highly training-efficient, shows sustained improvement, and achieves superior final generation quality.} This makes IAR2 highly effective and practical for large-scale image generative modeling.

\subsection{Exploration on Token Prediction Paradigms}
\label{sec:ablation_paradigm}

\hut{In this section and those that follow, we provide a detailed analysis of our proposed modules and hyperparameters. To ensure a consistent experimental setup, all experiments are conducted for 100 epochs with a fixed Classifier-Free Guidance (CFG) scale under 143M parameters (B version), unless specified otherwise.}

In this section, we conduct an ablation study to determine the optimal 
\yrexten{modeling strategy for autoregressively predicting} the dual-codebook token sequences. We design and evaluate four distinct architectural variants, each representing a different approach to handling the semantic ($k_i$) and detail ($j_i$) tokens. The configurations are detailed below, and their performance is summarized in Table~\ref{tab:paradigm_ablation}.

\begin{itemize}
    \item \textbf{Alternating Prediction:} This naive baseline processes a sequence of doubled length, $\{k_1, j_1, k_2, j_2, \ldots, k_m, j_m\}$, without token fusion. The model autoregressively predicts semantic and detail tokens in an alternating fashion, using a simple MLP head on the output hidden states.
    
    \item \textbf{Grouped Sequential Prediction:} Similar to the first baseline, this approach also operates on a doubled-length sequence but rearranges it as $\{k_1, \ldots, k_m, j_1, \ldots, j_m\}$. The model first predicts all semantic tokens for the entire image and then proceeds to predict all detail tokens. This method tests the hypothesis of separating the prediction process into two distinct stages.
    
    \item \textbf{Fused Independent Prediction:} This variant incorporates our token fusion mechanism (Eq.~\ref{eq:token compression}) to maintain the original sequence length. However, it employs two parallel MLP heads on the output hidden state $\hat{h}_i$ to predict $k_{i+1}$ and $j_{i+1}$ independently. This design overlooks the inherent conditional dependency of detail tokens on their semantic counterparts.
    
    \item \textbf{Fused Hierarchical Prediction (Ours):} Our proposed method utilizes token fusion for efficiency and employs our \textbf{Local-Context Enhanced AR Head} to perform hierarchical prediction. The model first predicts the semantic token $k_{i+1}$ from the hidden state $\hat{h}_i$, and then predicts the detail token $j_{i+1}$ conditioned on both $\hat{h}_i$ and the newly predicted $k_{i+1}$.
\end{itemize}

All models are trained for an equal amount of time, corresponding to the duration needed for our model to train for 100 epochs on ImageNet.
The results in Table~\ref{tab:paradigm_ablation} reveal several key insights. 
First, the \yrexten{two} baselines operating on doubled-length sequences (Alternating and Grouped Sequential \yrexten{Prediction}) yield suboptimal performance, likely due to the \yrexten{doubled sequence length,} increased computational complexity, and the challenge of modeling longer-range dependencies. 
\yrexten{Second}, the Fused Independent Prediction model performs the worst in terms of FID (7.92) and IS (168.61), which strongly validates our hypothesis that ignoring the semantic-to-detail dependency \yrexten{degrades the} generation quality. 
In contrast, our proposed Fused Hierarchical Prediction approach significantly outperforms all other variants, achieving the best FID (6.88), IS (175.70), and Recall (0.42). While the Grouped Sequential method achieves slightly higher precision, our model's superior recall indicates a much better ability to capture the diversity of the true data distribution. 
\textit{This study confirms that both token fusion (for efficiency) and \yrexten{the} hierarchical prediction mechanism (that \yrexten{considers} the \yrexten{inherent semantic-to-detail dependency} of visual data) are crucial for achieving state-of-the-art performance.}

\begin{table*}[t]
\centering
\caption{Exploration on different token prediction paradigms for the dual-codebook framework with (128, 4096) codebook size. Our approach demonstrates superior performance by efficiently modeling the semantic-to-detail dependency. Best results are in \textbf{bold}. Note that our model here has no local-context enhancement and progressive attention-guided CFG.}
\label{tab:paradigm_ablation}
\resizebox{0.98\textwidth}{!}{%
\begin{tabular}{@{}llcccc@{}}
\toprule
\textbf{Paradigm} & \textbf{Prediction Scheme} & \textbf{FID} $\downarrow$ & \textbf{IS} $\uparrow$ & \textbf{Precision} $\uparrow$ & \textbf{Recall} $\uparrow$ \\
\midrule
Alternating Prediction & Alternating $k_i, j_i$ prediction on a $2m$-length sequence & 7.22 & 173.17 & 0.85 & 0.37 \\
Grouped Sequential Prediction & Predict all $k$ tokens, then all $j$ tokens on a $2m$-length sequence & 7.48 & 172.00 & \textbf{0.86} & 0.36 \\
Fused Independent Prediction & Fused tokens; parallel MLP heads for independent $k_i, j_i$ prediction & 7.92 & 168.61 & 0.85 & 0.36 \\
\textbf{Fused Hierarchical Prediction (Ours)} & Fused tokens; AR head for hierarchical $k_i \to j_i$ prediction & \textbf{6.88} & \textbf{175.70} & 0.81 & \textbf{0.42} \\
\bottomrule
\end{tabular}%
}
\end{table*}

\begin{table}[t]
\caption{Ablation study on local-context enhancement and compression based on Fused Hierarchical Prediction in Table~\ref{tab:paradigm_ablation}.}
\label{tab:enhancement}
\centering
\resizebox{0.48\textwidth}{!}{ 
\begin{tabular}{cc|cccc}
\midrule
 Local Enh. & Compression & FID $\downarrow$ & IS $\uparrow$ & Prec. $\uparrow$ & Rec. $\uparrow$ \\
\midrule
  &  & 6.88 & 175.70 & 0.81 & \textbf{0.42} \\
 \checkmark &  & 6.66 & 180.84 & \textbf{0.85} & 0.39 \\
 \checkmark & \checkmark & \textbf{6.06} & \textbf{188.90} & 0.84 & \textbf{0.42} \\
\midrule
\end{tabular}}
\end{table}

\subsection{Exploration on the AR Head Compression}
\hut{To validate the efficacy of our proposed Local-Context Enhanced AR Head, and to investigate the role of the \yrexten{local-context enhancement and} context compression module \yrexten{within our AR head}, we conduct a detailed experiment on it. 
\yrexten{As} presented in Table~\ref{tab:enhancement}, \yrexten{we} compare three configurations built upon the same Fused Hierarchical Prediction backbone. To ensure a fair comparison, all models were trained for 
\yrexten{the same amount of time.}
}

\hut{First, we establish a baseline model that \yrexten{removes} local context enhancement \yrexten{and context compression} in the AR head. This model achieves an FID of 6.88 and an Inception Score (IS) of 175.70.
Next, we introduce local context \yrexten{enhancement} via naive concatenation of the full-dimensional hidden states from the local neighborhood, without using the compression module. This configuration shows an improvement over the baseline, reducing the FID to 6.66 and increasing the IS to 180.84, which confirms that incorporating local context is fundamentally beneficial. However, this approach incurs a significant computational overhead, which completes fewer training iterations within the fixed time budget. Consequently, its performance gain is limited.}

\hut{Finally, our full proposed model, which integrates both local context \yrexten{enhancement} and the lightweight compression module, achieves substantially superior performance. It obtains the best scores across the board, with a FID of \textbf{6.06} and an IS of \textbf{188.90}, while also restoring the recall to \textbf{0.42}. 
This highlights the dual advantage of our compression module. 
First, it drastically improves training efficiency, allowing the model to converge more effectively within the same training duration. Second, it distills the essential information from the local neighborhood into a compact and \yrexten{powerful} representation, enabling a more effective fusion with the global context. The significant performance leap validates that our compression strategy is \yrexten{crucial for making} 
the integration of local context both computationally feasible and maximally effective.}

\subsection{Exploration on Codebook Design}
\label{sec:ablation_codebook}

In this section, we investigate the impact of our proposed Semantic-Detail Associated Dual Codebook on generation quality. 
To \yrexten{validate} its contribution, we compare three distinct VQ-GAN architectures: (1) a standard baseline using a single codebook (Codebook from LLamaGen~\cite{sun2024llamagen} with codebook size 16384), (2) a model employing a dual codebook (Codebook size=(256,4096)) but without any explicit semantic-detail association~\cite{song2025dualtoken}, and (3) our proposed method, which structures the dual codebooks (Codebook size=(256,4096)) with a semantic-to-detail hierarchy. 
To ensure a fair comparison and isolate the contribution of the codebook design, the variant of our model used in this ablation does not employ the local-context enhancement or the progressive attention-guided CFG. All models are trained on the ImageNet~\cite{deng2009imagenet} dataset for 100 epochs.

As presented in Table~\ref{tab:codebook_ablation2}, the results offer a key insight into codebook design. Notably, a naive transition from a single-codebook architecture to \yrexten{an unassociated} dual-codebook setup leads to a performance degradation: the FID score increases from 6.60 to 6.74, while the Inception Score (IS) drops significantly from 187.2 to 169.2. This suggests that merely expanding representational capacity without a structured modeling framework introduces learning ambiguity and complicates the autoregressive task, ultimately harming generation quality. 
In contrast, our semantic-detail associated dual codebook not only reverses this negative trend but also surpasses the strong single-codebook baseline. It achieves a superior FID of 6.29 while restoring the IS to 186.3, demonstrating its ability to effectively harness the increased representational power for higher-fidelity synthesis. These findings empirically validate our core hypothesis: imposing a semantic-to-detail hierarchy \yrexten{with associations between dual codebooks} is crucial for unlocking the full potential of dual-codebook representations and achieving superior generative performance.

\begin{table}[t]
\centering

\caption{Comparison of different codebook architectures. Our proposed semantic-detail association is crucial for effectively leveraging a dual-codebook setup and outperforms both single-codebook and \yrexten{unassociated} dual-codebook baselines. Best results are in \textbf{bold}.}
\renewcommand{\arraystretch}{1.2}
\resizebox{0.48\textwidth}{!}{ 
\label{tab:codebook_ablation2}
\begin{tabular}{@{}lcccc@{}}
\toprule
\textbf{Method} & \textbf{FID} $\downarrow$ & \textbf{IS} $\uparrow$ & \textbf{Precision} $\uparrow$ & \textbf{Recall} $\uparrow$ \\
\midrule
Single Codebook  & 6.60 & \textbf{187.2} & 0.849 & 0.400 \\
\yrexten{Unassociated} Dual Codebook~\cite{song2025dualtoken} & 6.74 & 169.2 & 0.820 & \textbf{0.410} \\
\textbf{Associated Dual Codebook (Ours)} & \textbf{6.29} & 186.3 & \textbf{0.840} & \textbf{0.410} \\
\bottomrule
\end{tabular}}
\end{table}

\subsection{Exploration on Progressive Attention-Guided CFG}
\label{sec:ablation_cfg}

To validate the effectiveness of each component (\yrexten{attention-guided spatial modulation and progressive sequential scheduling}) within our PAG-CFG framework, we conduct an ablation study
as detailed in Table~\ref{tab:ablation on PAD-CFG}. The baseline model, employing a standard static CFG, establishes an FID of 5.13 after 300 epochs. Upon integrating the progressive schedule alone, we observe a clear improvement, with the FID decreasing to 4.95. This result confirms that dynamically strengthening the guidance throughout the generation process is effective for improving conditional alignment and overall sample fidelity.

The full PAG-CFG model, which combines the progressive schedule with attention guidance, achieves the best performance, further reducing the FID to a final score of \textbf{4.80}. This additional reduction demonstrates the crucial role of attention guidance. By spatially modulating the guidance strength, our method applies the intensified signal more precisely to semantically relevant regions, leading to an even greater enhancement in generation quality. The consistent improvement in FID scores across the configurations validates that both the progressive and attention-guided mechanisms are \yrexten{effective} and complementary, working together to achieve the optimal result.

\hutnew{To further substantiate the robustness of the attention-guided modulation, we evaluate its impact across various CFG scale configurations. As depicted in Figure~\ref{fig:cfg-compare}, we compare the generation performance with and without the attention guidance factor $\alpha$ under a range of start and end scale settings. The results consistently demonstrate that integrating the spatial modulation $\alpha$ yields superior image quality across all tested parameter combinations. This confirms that the attention guide provides a universal and consistent benefit, effectively enhancing the generative performance regardless of the specific progressive scheduling scales employed.}

\begin{table}[t]
\centering
\caption{Ablation study on the Progressive Attention-Guided CFG.}
\label{tab:ablation on PAD-CFG}
\resizebox{0.48\textwidth}{!}{%
\begin{tabular}{@{}cc|cccc|cccc@{}}
\toprule
\multirow{2}{*}{\makecell[c]{Progre-\\ssive}} & \multirow{2}{*}{\makecell[c]{Attn-\\Guide}}  & \multicolumn{4}{c|}{100 Epochs} & \multicolumn{4}{c}{300 Epochs} \\
\cmidrule(lr){3-6} \cmidrule(lr){7-10}
& & FID $\downarrow$ & IS $\uparrow$ & Prec. $\uparrow$ & Rec. $\uparrow$ & FID $\downarrow$ & IS $\uparrow$ & Prec. $\uparrow$ & Rec. $\uparrow$ \\
\midrule
  &  & 6.18 &187.9  & \textbf{0.85} & \textbf{0.40} &5.13 &\textbf{217.7} &\textbf{0.85} &0.43 \\
 \checkmark &  &6.03 &189.0 &\textbf{0.85} &\textbf{0.40} &4.95 &200.5 &0.83 &\textbf{0.46} \\
 \checkmark & \checkmark & \textbf{5.89} & {\textbf{192.8}}  & \textbf{0.85} & \textbf{0.40} & \textbf{4.80} & 211.8 & {0.84} & 0.45 \\
\bottomrule
\end{tabular}%
}
\end{table}

\subsection{\yrexten{More} Ablation Studies \yrexten{on Hyperparameters}}
To systematically verify the rationality of \yrexten{some} key hyperparameters, we design \yrexten{several} ablation experiments. All experiments are conducted on the ImageNet-256×256 dataset under a unified evaluation protocol.



\begin{table}[t]
\caption{Ablation study on the \yrexten{semantic loss} weight $\lambda_s$ \yrexten{in the loss function for training semantic-detail autoregressive prediction (Eq.~\ref{eq:loss_sd_ar_prediction})}. $\lambda_s$ balances the prediction of semantic tokens and detail tokens in the hierarchical autoregressive objective. Best results are in \textbf{bold}.}
\label{tab:abaltion study on loss weight}
\centering
\renewcommand{\arraystretch}{1.1}
\resizebox{0.33\textwidth}{!}{
\begin{tabular}{c|cccc}
\toprule
$\lambda_s$ & FID$\downarrow$ & IS$\uparrow$ & Precision$\uparrow$ & Recall$\uparrow$  \\
\midrule
0.5 &6.84&187.2&\textbf{0.86}&0.38 \\
1 &6.37&169.2&0.82&\textbf{0.43} \\
\textbf{2 (Ours)} &\textbf{6.18} &187.9&0.85&0.40\\
3 &6.23&\textbf{194.3}&0.85&0.40 \\
5 &6.30&192.7&0.84&0.40 \\
\bottomrule
\end{tabular}}
\end{table}

\begin{figure}[t]
\centering
\includegraphics[width=0.48\textwidth]{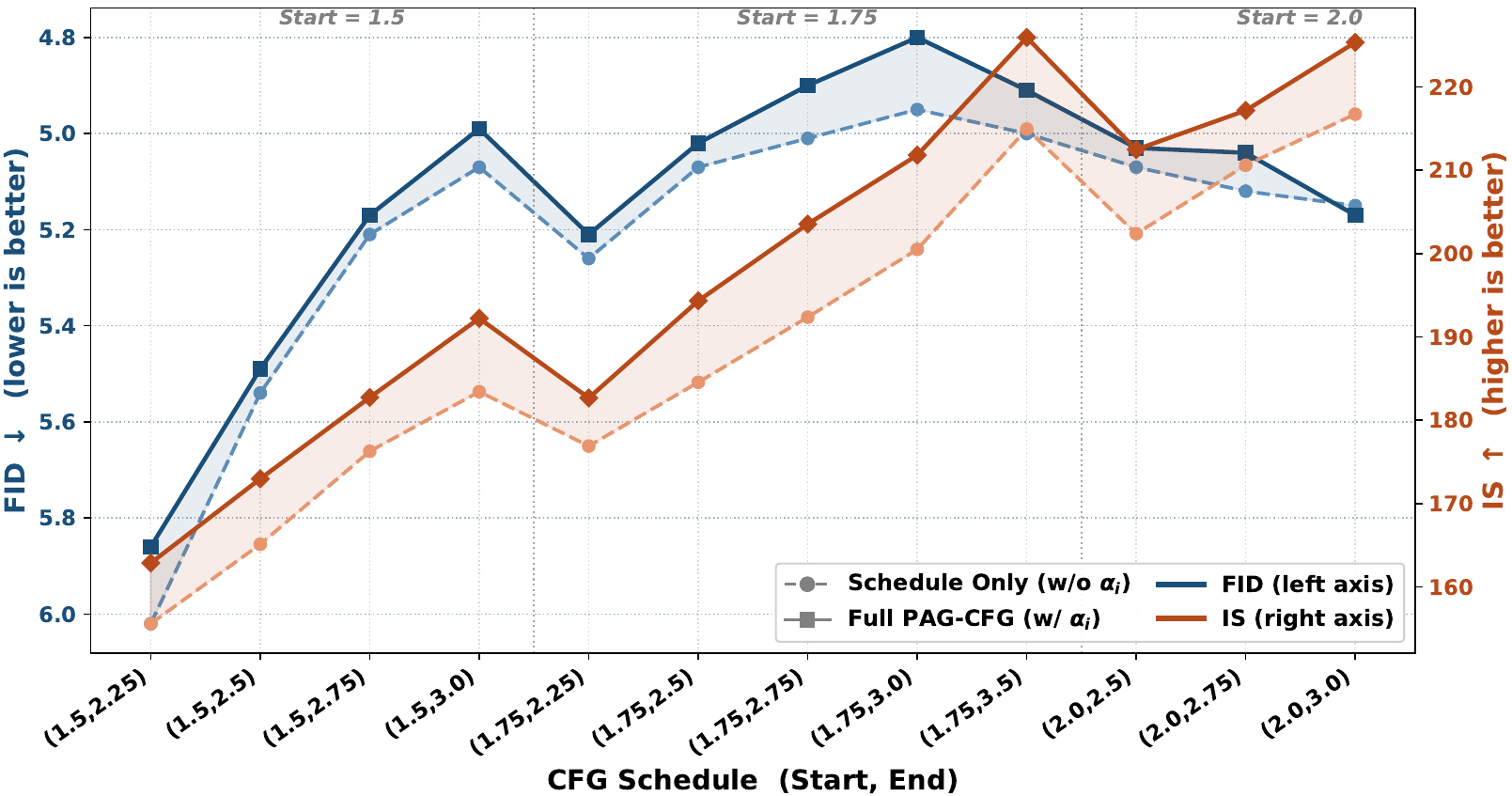}
\caption{\hutnew{Ablation study isolating the effect of Attention-Guided Modulation in PAG-CFG on IAR2-B. We compare the generation performance using \textit{only} the Progressive Schedule (setting $\alpha_i=1$, denoted as w/o $\alpha_i$) against the full PAG-CFG across various Start and End scale configurations. The comparison confirms that attention-guided spatial modulation consistently provides substantial improvements over a purely temporal schedule, regardless of the specific scaling bounds.}}
\label{fig:cfg-compare}
\end{figure}

\noindent\textbf{Ablation study on the hyperparameters in Progressive Attention-Guided CFG.}
We investigate the impact of the key hyperparameters in our PAG-CFG, namely the start\yrexten{ing guidance} scale $s_{start}$ and the end \yrexten{guidance} scale $s_{end}$, which together define the guidance schedule. We conduct a systematic grid search over a range of values, with the comprehensive results presented in Fig~\ref{fig:cfg-compare}. The findings reveal a clear and well-known trade-off between fidelity and diversity. Generally, increasing the guidance strength (either by raising $s_{start}$ or, more impactfully, $s_{end}$) leads to higher Inception Scores (IS) and Precision, indicating that the generated images are more \yrexten{diverse and more} distinctly recognizable 
as belonging to the target class. However, this enhanced alignment comes at the cost of a lower Recall score, suggesting a reduction in intra-class diversity.
Our primary goal is to find the configuration that optimizes overall image quality, for which FID is the most indicative metric. The results show that a starting scale of $s_{start}=1.75$ provides a superior \yrexten{FID} compared to $1.5$ or $2.0$. 
With $s_{start}$ fixed at $1.75$, we observe that the FID score consistently improves as $s_{end}$ increases, reaching its minimum (best) value of \textbf{4.80} when $s_{end}=3.0$. Although pushing $s_{end}$ further to $3.5$ achieves the highest IS (225.92) and a very high Precision (0.847), the FID score degrades to 4.91. This indicates that while the guidance becomes extremely effective at enforcing class \yrexten{alignment}, it begins to introduce artifacts that harm the overall realism of the images. Therefore, we select the setting ($s_{start}=1.75, s_{end}=3.0$) as our final configuration, as it strikes the most effective balance between achieving high fidelity (best FID), strong class-conditional alignment (high IS and Precision), and reasonable diversity.

\noindent\textbf{Ablation study on the semantic loss weight.}
\yrexten{In the loss function for training Semantic-Detail Autoregressive Prediction (Eq.~\ref{eq:loss_sd_ar_prediction}),} the semantic loss weight $\lambda_s$ serves as a critical hyperparameter to balance ``semantic accuracy’’ and \yrexten{``detail accuracy’’} in autoregressive generation. We \yrexten{conduct experiments varing} $\lambda_s$ from 0.5 to 5, with results summarized in Table~\ref{tab:abaltion study on loss weight}.
\hut{The findings reveal 
\yrexten{that:}
(i) An overly small $\lambda_s$ (e.g., 0.5) diminishes the penalty for incorrect semantic predictions. This can lead to an unstable semantic foundation, where the model generates details that are misaligned with the predicted content, thereby harming overall coherence.
(ii) Conversely, an excessively large $\lambda_s$ (e.g., 5) forces the model to prioritize semantic correctness at the expense of learning fine-grained details. While the high-level semantics might be correct, the generated images tend to lack \yrexten{details} and intricate textures, as the model is not sufficiently incentivized to predict detail tokens accurately.}
In summary, setting $\lambda_s=2$ strikes an effective balance between semantic-token and detail-token prediction, leading to the best overall generation quality.
\subsection{\hutnew{Computational Resource Analysis}}
\label{sec:comp_analysis}

To evaluate the computational efficiency of our Semantic-Detail Associated Dual Codebook framework, we compare the training and inference times of IAR2-B against a standard single-codebook baseline on a single GPU. The performance metrics, converted to seconds per iteration (s/iter), are summarized in Table~\ref{tab:computation_time}.

\begin{table}[h]
\centering
\caption{\hutnew{Computational time comparison between the single-codebook baseline and our dual-codebook IAR2-B on a single GPU.}}
\label{tab:computation_time}
\resizebox{0.48\textwidth}{!}{
\begin{tabular}{lcc}
\toprule
\textbf{Method} & \textbf{Training (s/iter)} $\downarrow$ & \textbf{Inference (s/iter)} $\downarrow$ \\
\midrule
Single-Codebook & 0.125 & 0.084 \\
IAR2-B (Ours) & 0.140 & 0.108 \\
\bottomrule
\end{tabular}
}
\end{table}

As shown in Table~\ref{tab:computation_time}, during the training phase, IAR2-B requires 0.140 s/iter compared to 0.125 s/iter for the baseline. Similarly, during the inference phase, IAR2-B takes 0.108 s/iter, whereas the single-codebook baseline requires 0.084 s/iter. This indicates that the dual-codebook design takes slightly more time per iteration in both phases. The increased computational cost is an expected consequence of our hierarchical prediction process, which requires the autoregressive model to decode both a semantic token and a detail token for each spatial patch, effectively expanding the sequence length and decoding steps.

However, this minor increase in per-iteration time is substantially outweighed by our model's superior convergence efficiency. As illustrated in Fig.~\ref{fig:more comparison}, the dual-codebook representation provides a more structured learning objective that significantly accelerates the optimization process. Empirically, IAR2-B achieves an acceleration of approximately 65\% in convergence compared to the single-codebook baseline. Consequently, despite taking slightly more time per iteration during training and inference, our approach effectively maximizes parameter updates, allowing it to achieve vastly superior generation quality within the same total wall-clock training time.

\section{Conclusion}
\label{sec:conclusion}

In this paper, we presented IAR2, an advanced autoregressive framework for image generation that addresses the limitations of prior methods, which often neglect the intrinsic structure of visual data. Building upon the insights from our previous work, IAR, but moving beyond its rigid codebook clustering, we introduced a \yrexten{hierarchical} semantic-detail synthesis process. This is enabled by three core contributions: the \textbf{Semantic-Detail Associated Dual Codebook} for a decoupled and more expressive representation, the \textbf{Local-Context Enhanced Autoregressive Head} for hierarchical and \yrexten{context}-aware prediction, and the \textbf{Progressive Attention-Guided Adaptive CFG} for dynamic conditional guidance. Together, these components create a \yrexten{cohesive} system that effectively \yrexten{achieves} global semantic coherence with fine-grained detail fidelity.
Our extensive experiments on the ImageNet benchmark validate the effectiveness of our approach. IAR2 establishes a new state-of-the-art, achieving a Fréchet Inception Distance (FID) of \textbf{1.50}. Notably, this result not only surpasses existing models in generation quality but also demonstrates superior computational efficiency, outperforming larger models trained with significantly more resources. The strong scaling properties observed further underscore the robustness and potential of our architecture, confirming that a structured approach to visual token modeling is a highly promising direction.

\section*{Data Availability}

The datasets analysed during the current study are publicly available in the ImageNet repository~\cite{deng2009imagenet}, which can be accessed at \url{https://www.image-net.org/}. No new datasets were generated during this study.

{\small
\bibliographystyle{unsrt}
\bibliography{ref}

@String(CVPR= {IEEE Conf. Comput. Vis. Pattern Recog.})

@String(ICCV= {Int. Conf. Comput. Vis.})

@String(ICLR = {Int. Conf. Learn. Represent.})

@String(AAAI = {AAAI})

@String(CVPR  = {CVPR})

@String(ICCV  = {ICCV})

@String(ICLR  = {ICLR})

@article{van2017vqvae,
  title={Neural discrete representation learning},
  author={Van Den Oord, Aaron and Vinyals, Oriol and others},
  journal={Advances in neural information processing systems},
  volume={30},
  year={2017}
}

@inproceedings{esser2021vqgan,
  title={Taming transformers for high-resolution image synthesis},
  author={Esser, Patrick and Rombach, Robin and Ommer, Bjorn},
  booktitle={Proceedings of the IEEE/CVF conference on computer vision and pattern recognition},
  pages={12873--12883},
  year={2021}
}

@article{yu2021ViT-VQGAN,
  title={Vector-quantized image modeling with improved vqgan},
  author={Yu, Jiahui and Li, Xin and Koh, Jing Yu and Zhang, Han and Pang, Ruoming and Qin, James and Ku, Alexander and Xu, Yuanzhong and Baldridge, Jason and Wu, Yonghui},
  journal={arXiv preprint arXiv:2110.04627},
  year={2021}
}

@inproceedings{lee2022rqvae,
  title={Autoregressive image generation using residual quantization},
  author={Lee, Doyup and Kim, Chiheon and Kim, Saehoon and Cho, Minsu and Han, Wook-Shin},
  booktitle={Proceedings of the IEEE/CVF conference on computer vision and pattern recognition},
  pages={11523--11532},
  year={2022}
}

@inproceedings{npp,
  title={Next patch prediction for autoregressive visual generation},
  author={Pang, Yatian and Jin, Peng and Yang, Shuo and Zhu, Bin and Lin, Bin and Feng, Chaoran and Tang, Zhenyu and Chen, Liuhan and Tay, Francis EH and Lim, Ser-Nam and others},
  booktitle={Proceedings of the AAAI Conference on Artificial Intelligence},
  volume={40},
  number={10},
  pages={8260--8268},
  year={2026}
}

@article{ma2026unitok,
  title={Unitok: A unified tokenizer for visual generation and understanding},
  author={Ma, Chuofan and Jiang, Yi and Wu, Junfeng and Yang, Jihan and Yu, Xin and Yuan, Zehuan and Peng, Bingyue and Qi, Xiaojuan},
  journal={Advances in Neural Information Processing Systems},
  volume={38},
  pages={129274--129297},
  year={2026}
}

@inproceedings{wang2025parallelized,
  title={Parallelized autoregressive visual generation},
  author={Wang, Yuqing and Ren, Shuhuai and Lin, Zhijie and Han, Yujin and Guo, Haoyuan and Yang, Zhenheng and Zou, Difan and Feng, Jiashi and Liu, Xihui},
  booktitle={Proceedings of the IEEE/CVF Conference on Computer Vision and Pattern Recognition},
  pages={12955--12965},
  year={2025}
}

@article{bai2024fqgan,
  title={Factorized visual tokenization and generation},
  author={Bai, Zechen and Gao, Jianxiong and Gao, Ziteng and Wang, Pichao and Zhang, Zheng and He, Tong and Shou, Mike Zheng},
  journal={arXiv preprint arXiv:2411.16681},
  year={2024}
}

@inproceedings{qu2025tokenflow,
  title={Tokenflow: Unified image tokenizer for multimodal understanding and generation},
  author={Qu, Liao and Zhang, Huichao and Liu, Yiheng and Wang, Xu and Jiang, Yi and Gao, Yiming and Ye, Hu and Du, Daniel K and Yuan, Zehuan and Wu, Xinglong},
  booktitle={Proceedings of the Computer Vision and Pattern Recognition Conference},
  pages={2545--2555},
  year={2025}
}

@article{yu2023MAGVIT-v2,
  title={Language Model Beats Diffusion--Tokenizer is Key to Visual Generation},
  author={Yu, Lijun and Lezama, Jos{\'e} and Gundavarapu, Nitesh B and Versari, Luca and Sohn, Kihyuk and Minnen, David and Cheng, Yong and Birodkar, Vighnesh and Gupta, Agrim and Gu, Xiuye and others},
  journal={arXiv preprint arXiv:2310.05737},
  year={2023}
}

@article{luo2024Open-magvit2,
  title={Open-magvit2: An open-source project toward democratizing auto-regressive visual generation},
  author={Luo, Zhuoyan and Shi, Fengyuan and Ge, Yixiao and Yang, Yujiu and Wang, Limin and Shan, Ying},
  journal={arXiv preprint arXiv:2409.04410},
  year={2024}
}

@article{song2025dualtoken,
  title={Dualtoken: Towards unifying visual understanding and generation with dual visual vocabularies},
  author={Song, Wei and Wang, Yuran and Song, Zijia and Li, Yadong and Sun, Haoze and Chen, Weipeng and Zhou, Zenan and Xu, Jianhua and Wang, Jiaqi and Yu, Kaicheng},
  journal={arXiv preprint arXiv:2503.14324},
  year={2025}
}

@inproceedings{karras2019stylegan,
  title={A style-based generator architecture for generative adversarial networks},
  author={Karras, Tero and Laine, Samuli and Aila, Timo},
  booktitle={Proceedings of the IEEE/CVF conference on computer vision and pattern recognition},
  pages={4401--4410},
  year={2019}
}

@article{ho2020ddpm,
  title={Denoising diffusion probabilistic models},
  author={Ho, Jonathan and Jain, Ajay and Abbeel, Pieter},
  journal={Advances in neural information processing systems},
  volume={33},
  pages={6840--6851},
  year={2020}
}

@inproceedings{nichol2021iddpm,
  title={Improved denoising diffusion probabilistic models},
  author={Nichol, Alexander Quinn and Dhariwal, Prafulla},
  booktitle={International conference on machine learning},
  pages={8162--8171},
  year={2021},
  organization={PMLR}
}

@article{song2020ddim,
  title={Denoising diffusion implicit models},
  author={Song, Jiaming and Meng, Chenlin and Ermon, Stefano},
  journal={arXiv preprint arXiv:2010.02502},
  year={2020}
}

@article{dhariwal2021adm,
  title={Diffusion models beat gans on image synthesis},
  author={Dhariwal, Prafulla and Nichol, Alexander},
  journal={Advances in neural information processing systems},
  volume={34},
  pages={8780--8794},
  year={2021}
}

@article{saharia2022Imagen,
  title={Photorealistic text-to-image diffusion models with deep language understanding},
  author={Saharia, Chitwan and Chan, William and Saxena, Saurabh and Li, Lala and Whang, Jay and Denton, Emily L and Ghasemipour, Kamyar and Gontijo Lopes, Raphael and Karagol Ayan, Burcu and Salimans, Tim and others},
  journal={Advances in neural information processing systems},
  volume={35},
  pages={36479--36494},
  year={2022}
}

@inproceedings{rombach2022stable,
  title={High-resolution image synthesis with latent diffusion models},
  author={Rombach, Robin and Blattmann, Andreas and Lorenz, Dominik and Esser, Patrick and Ommer, Bj{\"o}rn},
  booktitle={Proceedings of the IEEE/CVF conference on computer vision and pattern recognition},
  pages={10684--10695},
  year={2022}
}

@inproceedings{ramesh2021DALLE,
  title={Zero-shot text-to-image generation},
  author={Ramesh, Aditya and Pavlov, Mikhail and Goh, Gabriel and Gray, Scott and Voss, Chelsea and Radford, Alec and Chen, Mark and Sutskever, Ilya},
  booktitle={International conference on machine learning},
  pages={8821--8831},
  year={2021},
  organization={Pmlr}
}

@article{yu2022Parti,
  title={Scaling autoregressive models for content-rich text-to-image generation},
  author={Yu, Jiahui and Xu, Yuanzhong and Koh, Jing Yu and Luong, Thang and Baid, Gunjan and Wang, Zirui and Vasudevan, Vijay and Ku, Alexander and Yang, Yinfei and Ayan, Burcu Karagol and others},
  journal={arXiv preprint arXiv:2206.10789},
  volume={2},
  number={3},
  pages={5},
  year={2022}
}

@article{sun2024llamagen,
  title={Autoregressive model beats diffusion: Llama for scalable image generation},
  author={Sun, Peize and Jiang, Yi and Chen, Shoufa and Zhang, Shilong and Peng, Bingyue and Luo, Ping and Yuan, Zehuan},
  journal={arXiv preprint arXiv:2406.06525},
  year={2024}
}

@inproceedings{hu2025iar,
  title={Improving autoregressive visual generation with cluster-oriented token prediction},
  author={Hu, Teng and Zhang, Jiangning and Yi, Ran and Weng, Jieyu and Wang, Yabiao and Zeng, Xianfang and Xue, Zhucun and Ma, Lizhuang},
  booktitle={Proceedings of the Computer Vision and Pattern Recognition Conference},
  pages={9351--9360},
  year={2025}
}

@inproceedings{chang2022maskgit,
  title={Maskgit: Masked generative image transformer},
  author={Chang, Huiwen and Zhang, Han and Jiang, Lu and Liu, Ce and Freeman, William T},
  booktitle={Proceedings of the IEEE/CVF conference on computer vision and pattern recognition},
  pages={11315--11325},
  year={2022}
}

@inproceedings{yu2023magvit,
  title={Magvit: Masked generative video transformer},
  author={Yu, Lijun and Cheng, Yong and Sohn, Kihyuk and Lezama, Jos{\'e} and Zhang, Han and Chang, Huiwen and Hauptmann, Alexander G and Yang, Ming-Hsuan and Hao, Yuan and Essa, Irfan and others},
  booktitle={Proceedings of the IEEE/CVF Conference on Computer Vision and Pattern Recognition},
  pages={10459--10469},
  year={2023}
}

@article{chang2023muse,
  title={Muse: Text-to-image generation via masked generative transformers},
  author={Chang, Huiwen and Zhang, Han and Barber, Jarred and Maschinot, AJ and Lezama, Jose and Jiang, Lu and Yang, Ming-Hsuan and Murphy, Kevin and Freeman, William T and Rubinstein, Michael and others},
  journal={arXiv preprint arXiv:2301.00704},
  year={2023}
}

@article{mirza2014cgan,
  title={Conditional generative adversarial nets},
  author={Mirza, Mehdi and Osindero, Simon},
  journal={arXiv preprint arXiv:1411.1784},
  year={2014}
}

@article{radford2015dcgan,
  title={Unsupervised representation learning with deep convolutional generative adversarial networks},
  author={Radford, Alec and Metz, Luke and Chintala, Soumith},
  journal={arXiv preprint arXiv:1511.06434},
  year={2015}
}

@article{salimans2016improvedgans,
  title={Improved techniques for training gans},
  author={Salimans, Tim and Goodfellow, Ian and Zaremba, Wojciech and Cheung, Vicki and Radford, Alec and Chen, Xi},
  journal={Advances in neural information processing systems},
  volume={29},
  year={2016}
}

@inproceedings{isola2017Pix2Pix,
  title={Image-to-image translation with conditional adversarial networks},
  author={Isola, Phillip and Zhu, Jun-Yan and Zhou, Tinghui and Efros, Alexei A},
  booktitle={Proceedings of the IEEE conference on computer vision and pattern recognition},
  pages={1125--1134},
  year={2017}
}

@inproceedings{zhu2017CycleGAN,
  title={Unpaired image-to-image translation using cycle-consistent adversarial networks},
  author={Zhu, Jun-Yan and Park, Taesung and Isola, Phillip and Efros, Alexei A},
  booktitle={Proceedings of the IEEE international conference on computer vision},
  pages={2223--2232},
  year={2017}
}

@article{goodfellow2020gan,
  title={Generative adversarial networks},
  author={Goodfellow, Ian and Pouget-Abadie, Jean and Mirza, Mehdi and Xu, Bing and Warde-Farley, David and Ozair, Sherjil and Courville, Aaron and Bengio, Yoshua},
  journal={Commun Acm},
  //volume={63},
  //number={11},
  //pages={139--144},
  year={2020},
  //publisher={ACM New York, NY, USA}
}

@inproceedings{ddpm,
  title={Denoising diffusion probabilistic models},
  author={Ho, Jonathan and Jain, Ajay and Abbeel, Pieter},
  booktitle={NeurIPS},
  //volume={33},
  //pages={6840--6851},
  year={2020}
}

@inproceedings{VAR,
  title={Visual autoregressive modeling: Scalable image generation via next-scale prediction},
  author={Tian, Keyu and Jiang, Yi and Yuan, Zehuan and Peng, Bingyue and Wang, Liwei},
  booktitle={NeurIPS},
  year={2024}
}

@inproceedings{maskgit,
  title={Maskgit: Masked generative image transformer},
  author={Chang, Huiwen and Zhang, Han and Jiang, Lu and Liu, Ce and Freeman, William T},
  booktitle={CVPR},
  //pages={11315--11325},
  year={2022}
}

@inproceedings{vqgan,
  title={Taming transformers for high-resolution image synthesis},
  author={Esser, Patrick and Rombach, Robin and Ommer, Bjorn},
  booktitle={CVPR},
  //pages={12873--12883},
  year={2021}
}

@article{meissonic,
  title={Meissonic: Revitalizing Masked Generative Transformers for Efficient High-Resolution Text-to-Image Synthesis},
  author={Bai, Jinbin and Ye, Tian and Chow, Wei and Song, Enxin and Chen, Qing-Guo and Li, Xiangtai and Dong, Zhen and Zhu, Lei and Yan, Shuicheng},
  journal={arXiv preprint arXiv:2410.08261},
  year={2024}
}

@article{llama,
  title={Llama: Open and efficient foundation language models},
  author={Touvron, Hugo and Lavril, Thibaut and Izacard, Gautier and Martinet, Xavier and Lachaux, Marie-Anne and Lacroix, Timoth{\'e}e and Rozi{\`e}re, Baptiste and Goyal, Naman and Hambro, Eric and Azhar, Faisal and others},
  journal={arXiv preprint arXiv:2302.13971},
  year={2023}
}

@article{gpt,
  title={Language models are unsupervised multitask learners},
  author={Radford, Alec and Wu, Jeffrey and Child, Rewon and Luan, David and Amodei, Dario and Sutskever, Ilya and others},
  journal={OpenAI blog},
  //volume={1},
  //number={8},
  //pages={9},
  year={2019}
}

@article{bert,
  title={Bert: Pre-training of deep bidirectional transformers for language understanding},
  author={Devlin, Jacob},
  journal={arXiv preprint arXiv:1810.04805},
  year={2018}
}

@article{gpt4,
  title={Gpt-4 technical report},
  author={Achiam, Josh and Adler, Steven and Agarwal, Sandhini and Ahmad, Lama and Akkaya, Ilge and Aleman, Florencia Leoni and Almeida, Diogo and Altenschmidt, Janko and Altman, Sam and Anadkat, Shyamal and others},
  journal={arXiv preprint arXiv:2303.08774},
  year={2023}
}

@article{cdm,
  title={Cascaded diffusion models for high fidelity image generation},
  author={Ho, Jonathan and Saharia, Chitwan and Chan, William and Fleet, David J and Norouzi, Mohammad and Salimans, Tim},
  journal={JMLR},
  //volume={23},
  //number={1},
  //pages={2249--2281},
  year={2022},
  //publisher={JMLRORG}
}

@inproceedings{ldm,
  title={High-resolution image synthesis with latent diffusion models},
  author={Rombach, Robin and Blattmann, Andreas and Lorenz, Dominik and Esser, Patrick and Ommer, Bj{\"o}rn},
  booktitle={CVPR},
  //pages={10684--10695},
  year={2022}
}

@inproceedings{dit,
  title={Scalable diffusion models with transformers},
  author={Peebles, William and Xie, Saining},
  booktitle={ICCV},
  //pages={4195--4205},
  year={2023}
}

@inproceedings{stylegan-xl,
  title={Stylegan-xl: Scaling stylegan to large diverse datasets},
  author={Sauer, Axel and Schwarz, Katja and Geiger, Andreas},
  booktitle={ACM SIGGRAPH},
  //pages={1--10},
  year={2022}
}

@inproceedings{gigagan,
  title={Scaling up gans for text-to-image synthesis},
  author={Kang, Minguk and Zhu, Jun-Yan and Zhang, Richard and Park, Jaesik and Shechtman, Eli and Paris, Sylvain and Park, Taesung},
  booktitle={CVPR},
  //pages={10124--10134},
  year={2023}
}

@article{biggan,
  title={Large scale GAN training for high fidelity natural image synthesis},
  author={Brock, Andrew and Donahue, Jeff and Simonyan, Karen},
  journal={arXiv preprint arXiv:1809.11096},
  year={2018}
}

@inproceedings{lpips,
  title={The unreasonable effectiveness of deep features as a perceptual metric},
  author={Zhang, Richard and Isola, Phillip and Efros, Alexei A and Shechtman, Eli and Wang, Oliver},
  booktitle={CVPR},
  //pages={586--595},
  year={2018}
}

@inproceedings{fid,
  title={Gans trained by a two time-scale update rule converge to a local nash equilibrium},
  author={Heusel, Martin and Ramsauer, Hubert and Unterthiner, Thomas and Nessler, Bernhard and Hochreiter, Sepp},
  booktitle={NeurIPS},
  //volume={30},
  year={2017}
}

@inproceedings{inception_score,
  title={Improved techniques for training gans},
  author={Salimans, Tim and Goodfellow, Ian and Zaremba, Wojciech and Cheung, Vicki and Radford, Alec and Chen, Xi},
  booktitle={NeurIPS},
  //volume={29},
  year={2016}
}

@inproceedings{vit-vqgan,
  title={Vector-quantized image modeling with improved vqgan},
  author={Yu, Jiahui and Li, Xin and Koh, Jing Yu and Zhang, Han and Pang, Ruoming and Qin, James and Ku, Alexander and Xu, Yuanzhong and Baldridge, Jason and Wu, Yonghui},
  booktitle={ICLR},
  year={2021}
}

@inproceedings{rq,
  title={Autoregressive image generation using residual quantization},
  author={Lee, Doyup and Kim, Chiheon and Kim, Saehoon and Cho, Minsu and Han, Wook-Shin},
  booktitle={CVPR},
  //pages={11523--11532},
  year={2022}
}

@inproceedings{adm,
  title={Diffusion models beat gans on image synthesis},
  author={Dhariwal, Prafulla and Nichol, Alexander},
  booktitle={NeurIPS},
  //volume={34},
  //pages={8780--8794},
  year={2021}
}

@inproceedings{deng2009imagenet,
  title={Imagenet: A large-scale hierarchical image database},
  author={Deng, Jia and Dong, Wei and Socher, Richard and Li, Li-Jia and Li, Kai and Fei-Fei, Li},
  booktitle={CVPR},
  //pages={248--255},
  year={2009},
  //organization={Ieee}
}

@inproceedings{precision_and_recall,
  title={Improved precision and recall metric for assessing generative models},
  author={Kynk{\"a}{\"a}nniemi, Tuomas and Karras, Tero and Laine, Samuli and Lehtinen, Jaakko and Aila, Timo},
  booktitle={NeurIPS},
  //volume={32},
  year={2019}
}

@article{cfg,
  title={Classifier-free diffusion guidance},
  author={Ho, Jonathan and Salimans, Tim},
  journal={arXiv preprint arXiv:2207.12598},
  year={2022}
}

@article{topk,
  title={Mining top-K frequent itemsets through progressive sampling},
  author={Pietracaprina, Andrea and Riondato, Matteo and Upfal, Eli and Vandin, Fabio},
  journal={DATAMINE},
  //volume={21},
  //pages={310--326},
  year={2010},
  //publisher={Springer}
}

@inproceedings{topp,
  title={The curious case of neural text degeneration},
  author={Holtzman, Ari and Buys, Jan and Du, Li and Forbes, Maxwell and Choi, Yejin},
  booktitle={ICLR},
  year={2019}
}

@article{temperature,
    author = {David H. Ackley and Geoffrey E. Hinton and Terrence J. Sejnowski},
    title = {A learning algorithm for boltzmann machines},
    journal = {Cognitive Science},
    year = {1985}
}

@inproceedings{temperature2,
  title={Synthetic literature: Writing science fiction in a co-creative process},
  author={Manjavacas, Enrique and Karsdorp, Folgert and Burtenshaw, Ben and Kestemont, Mike},
  booktitle={CCNLG},
  //pages={29--37},
  year={2017}
}
}

\appendix

\section{Overview}

In this supplementary material, more details about the proposed IAR2 method and more experimental results are provided, including:

\begin{itemize}
    \item More implementation details (Sec.~\ref{sec:more implementation results});
    \item More comparisons on the codebook reconstruction capability (Sec.~\ref{sec:Codebook reconstruction comparison});
    \item More comparisons on the training loss under different model sizes (Sec.~\ref{sec:Training loss});
    \item Further Analysis of the Semantic-Detail Associated Dual Codebook (Sec.~\ref{sec:supp_dual_codebook})
    \item More visualization resutls (Sec.~\ref{sec:more visualization results}).
\end{itemize}
\yrexten{The source code of IAR2 is available at: \url{https://github.com/sjtuplayer/IAR2}}.

\section{More Implementation Details}
\label{sec:more implementation results}

\textbf{Experimental Setup.}
Our experimental setup adheres to the protocol established by LlamaGen~\cite{sun2024llamagen}, ensuring consistency in hyperparameters for fair comparison. Detailed configurations for the training and inference phases are provided in Table~\ref{tab:training settings} and Table~\ref{tab:inference settings}, respectively.

\begin{table*}[h]
\caption{The training settings and hyperparameters used in our model. ``Const.'' denotes constant learning rate, while ``Cosine'' denotes cosine decay from $1.5\times10^{-4}$ to $5\times10^{-5}$. $\lambda_s$ is the weight for semantic cross-entropy loss.}
\label{tab:training settings}
\centering
\resizebox{1.0\textwidth}{!}{
\begin{tabular}{l|*{4}{>{\centering\arraybackslash}p{1.7cm}}|*{4}{>{\centering\arraybackslash}p{1.7cm}}}
\toprule
Model                  & B&L&XL&XXL           & B&L&XL&XXL          \\ \midrule
Parameter Num          & 143M     & 408M     & 884M     & 1.5B     & 143M     & 408M     & 884M     & 1.5B     \\ \midrule
Token Num              & \multicolumn{4}{c|}{16$\times$16}                 & \multicolumn{4}{c}{24$\times$24}                 \\ \midrule

Optimizer              & \multicolumn{8}{c}{AdamW}                                                             \\
Weight decay           & \multicolumn{8}{c}{0.05}                                                              \\ \midrule
Batch Size             & 256      & 256      & 256      & 256      & 256      & 256      & 256      & 512      \\
Learning Rate          & 1e-4 & 1e-4 & 1.5e-4$\rightarrow$5e-5 & 1.5e-4$\rightarrow$5e-5 
             & 1e-4 & 1e-4 & 1.5e-4$\rightarrow$5e-5 &3e-4$\rightarrow$1e-4 \\
LR Scheduler           & Const. & Const. & Cosine & Cosine & Const. & Const. & Cosine & Cosine \\
GPU Num                &    16    &    16    &    8   &    8    & 16       & 16      & 16        & 32       \\
Epoch                  & 300      & 300      & 50       & 50       & 300      & 300      & 300      & 300     \\
FSDP                   & Yes & Yes & No & No & Yes & Yes & No & Yes \\ 
$\lambda_s$ & 2.0 & 1.0 & 1.0 & 1.0 & 2.0 & 1.0 & 1.0 & 1.5 \\ 
\bottomrule
\end{tabular}}
\end{table*}

\begin{table*}[h]
\caption{The inference settings and hyperparameters used in the experiments..}
\label{tab:inference settings}
\centering
\resizebox{1.0\textwidth}{!}{
\begin{tabular}{l|*{4}{>{\centering\arraybackslash}p{1.7cm}}|*{4}{>{\centering\arraybackslash}p{1.7cm}}}
\toprule
Model                  & B&L&XL&XXL           & B&L&XL&XXL          \\ \midrule
Parameter Num          & 143M     & 408M     & 884M     & 1.5B     & 143M     & 408M     & 884M     & 1.5B     \\ \midrule
Token Num              & \multicolumn{4}{c|}{16$\times$16}                 & \multicolumn{4}{c}{24$\times$24}                 \\ \midrule
Random Seed   & \multicolumn{8}{c}{0}                                 \\
Top K         & \multicolumn{8}{c}{0}                                 \\
Top P         & \multicolumn{8}{c}{1.0}                                 \\
Temperature   & \multicolumn{8}{c}{1.0}                                 \\
 \midrule
CFG           & 1.75$\rightarrow$2.5  &  1.4$\rightarrow$2.5 & 1.25$\rightarrow$3.0 &1.25$\rightarrow$3.0 & 1.75$\rightarrow$3.0 & 1.4$\rightarrow$2.5 &  1.35$\rightarrow$3.0 &1.4$\rightarrow$3.15  \\
\bottomrule
\end{tabular}}
\end{table*}

\noindent\textbf{Sampling Strategies and Hyperparameters.}
During the inference phase, several key hyperparameters and sampling strategies are employed to control the generation process. We detail these below:

\begin{itemize}

    \item \textbf{Top-K Sampling:} This decoding strategy~\cite{topk} restricts the sampling space to the $k$ most probable tokens at each step. While this method focuses on high-probability candidates, its fixed vocabulary size ($k$) can sometimes prematurely discard viable, lower-probability tokens.

    \item \textbf{Top-P (Nucleus) Sampling:} Alternatively, Top-P sampling~\cite{topp}, also known as nucleus sampling, dynamically constructs a candidate set by selecting the smallest group of tokens whose cumulative probability mass is at least $p$. This adaptive approach tailors the sampling vocabulary to the local probability distribution, effectively balancing coherence and diversity in the generated sequence.

    \item \textbf{Temperature Scaling:} The temperature hyperparameter~\cite{temperature,temperature2} modulates the randomness of the sampling process by rescaling the logit values before the softmax operation. A lower temperature ($T < 1$) sharpens the distribution, making the model's output more deterministic. Conversely, a higher temperature ($T > 1$) flattens the distribution, promoting diversity. The rescaled probability $P_i$ for the $i$-th token is computed as:
    \[
    P_i = \frac{\exp{(l_i / T)}}{\sum_j \exp{(l_j / T)}},
    \]
    where $l_i$ is the logit for the $i$-th token and $T$ is the temperature.
\end{itemize}

\section{Codebook Reconstruction Comparison}
\label{sec:Codebook reconstruction comparison}
In this section, we evaluate the fidelity of the codebook reconstruction capability of our learned image tokenizer against several prominent methods, namely VQGAN~\cite{vqgan}, MaskGIT~\cite{maskgit}, and LlamaGen~\cite{sun2024llamagen}. This comparison is crucial as a high-fidelity tokenizer is a prerequisite for achieving superior results in subsequent autoregressive generation tasks.

The quantitative results are summarized in Table \ref{Codebook Reconstruction Comparison}. As shown, our method achieves a notably superior reconstruction performance across all metrics. Specifically, our tokenizer \yrexten{achieves} an rFID \yrexten{(reconstruction FID)} of 1.05, which is a significant improvement over the next best-performing methods, LlamaGen (2.19) and MaskGIT (2.28). This low rFID suggests that the images reconstructed from our codebook are perceptually much closer to the original inputs.

Furthermore, our \yrexten{IAR2} model yields the highest pixel-level fidelity, with a PSNR of 21.71 and an SSIM of 0.702. This performance is achieved with a relatively small codebook size of 4352 entries \yrexten{(256 for semantic, 4096 for detail codebook)} and a compact latent dimension of 8, demonstrating a highly efficient and effective representation learning. In contrast, while VQGAN models can achieve reasonable performance, they require a much larger latent dimension (256) and a larger codebook size (16384), yet still fall short of our model's reconstruction quality (e.g., VQGAN 16384 achieves rFID of 4.99 and PSNR of 20.00).

The results decisively establish the effectiveness of our proposed tokenizer architecture in learning a discretized latent space that preserves critical image information while simultaneously offering a compact and high-fidelity representation suitable for subsequent autoregressive modeling.

\begin{table}[t]
\caption{Comparisons with other image tokenizers. The evaluations are on 256×256 ImageNet 50k validation set, with a downsampling rate of 16.}
\label{Codebook Reconstruction Comparison}
\centering
\resizebox{0.48\textwidth}{!}{
\begin{tabular}{lcc|ccc}
\toprule
Method & dim & size & rFID$\downarrow$ & {PSNR$\uparrow$} & SSIM$\uparrow$ \\
\midrule
VQGAN & 256 & 1024 & 8.30 & 19.51 & 0.614 \\
VQGAN & 256 & 16384 & 4.99 & 20.00 & 0.629 \\
MaskGIT & 256 & 1024 & 2.28 & - & - \\
LlamaGen & 8 & 16384 & 2.19 & 20.79 & 0.675 \\
\textbf{IAR2 (Ours)} & 8 & (256,4096) & \textbf{1.05} & \textbf{21.71} & \textbf{0.702} \\
\bottomrule
\end{tabular}}
\end{table}

\begin{figure}[t]
\centering
\includegraphics[width=0.48\textwidth]{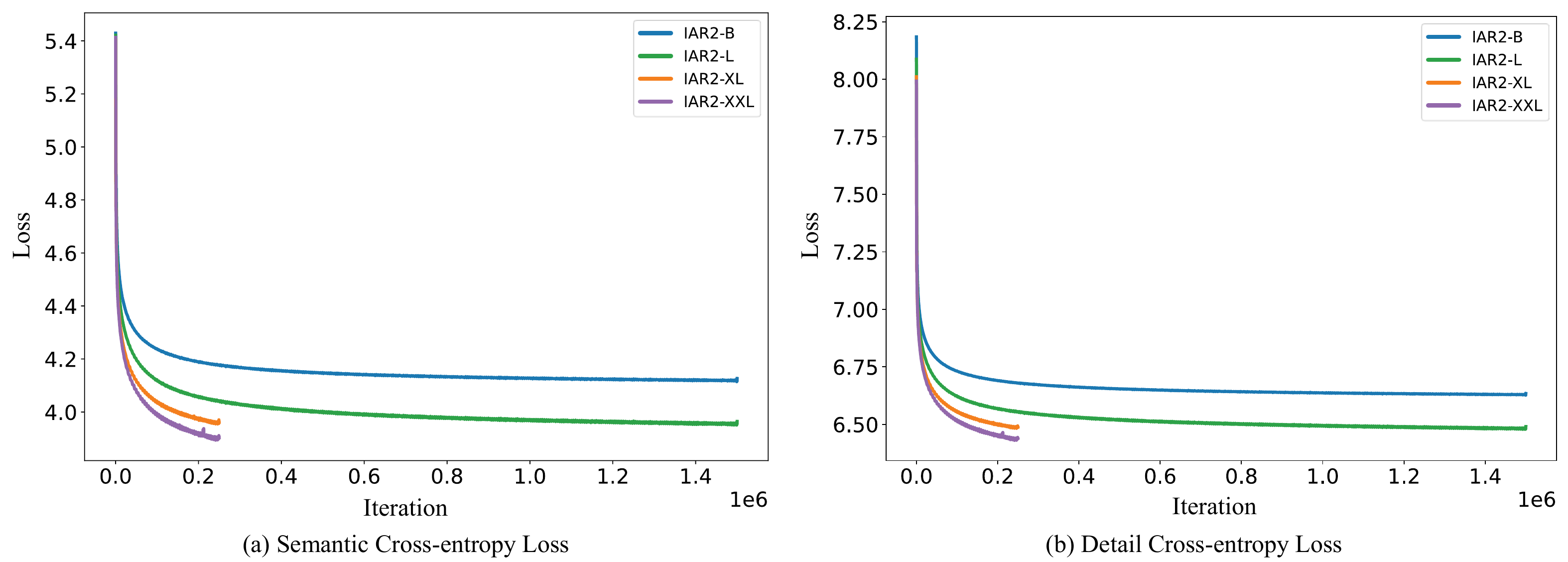}
\caption{The training loss curves for the semantic cross-entropy loss (a) and the detail cross-entropy loss (b) on $\mathbf{16\times 16}$ \textbf{image tokens}.}
\label{fig: loss-curves-384}
\end{figure}

\section{Training Losses Under Different Model Sizes}
\label{sec:Training loss}

Figure~\ref{fig: loss-curves-384} illustrates the training loss curves for models of varying sizes throughout the training process (with $\mathbf{16\times 16}$ image tokens). As shown, larger models consistently achieve lower loss values across iterations compared to their smaller counterparts. This observation validates the scaling capability of our model architecture: with more parameters, the model is able to better capture the underlying data distribution and fit the training set more effectively. Notably, the largest model (IAR2-XXL) achieves the fastest convergence and the lowest final loss, indicating enhanced optimization efficiency as well as increased representational power. These findings suggest that scaling up the model contributes positively to its training dynamics, supporting the efficacy of our approach for accommodating larger and more complex datasets.

\section{\hutnew{Further Analysis of the Semantic-Detail Associated Dual Codebook}}
\label{sec:supp_dual_codebook}

To address the concerns regarding the distinct functions and the necessity of coupling in our Semantic-Detail Associated Dual Codebook (discussed in Section 4.2 and Table 7 of the main text), we provide further visual and quantitative analyses.

\subsection{\hutnew{Visual and Quantitative Analysis of Distinct Functions}}
Although our dual-codebook design expands the representational capacity to a polynomial scale ($n_1 \times n_2$), it is crucial to understand what each codebook actually learns. To demonstrate their distinct functions, we provide both quantitative reconstruction metrics and visual comparisons using different combinations of the codebooks.

Quantitatively, as shown in Table~\ref{tab:supp_recon}, relying solely on the Semantic Codebook (size 256) yields a reconstruction FID (rFID) of 11.35. This indicates that while it captures basic global semantics, it loses significant high-frequency details. Using only the Detail Codebook (size 4096) improves the rFID to 5.77, but it is still fundamentally bottlenecked by the limited linear vocabulary capacity. In contrast, our Associated Dual Codebook (combining both codebooks) dramatically improves the reconstruction quality, achieving an rFID of 1.72, a PSNR of 20.95, and an SSIM of 0.67. This substantial leap in performance confirms that the two codebooks learn highly complementary information rather than redundant features.

\begin{table}[h]
\centering
\caption{Quantitative comparison of reconstruction performance using different codebook configurations. The Associated Dual Codebook leverages the complementary strengths of both codebooks to achieve significantly superior reconstruction quality.}
\renewcommand{\arraystretch}{1.2}
\resizebox{0.48\textwidth}{!}{
\label{tab:supp_recon}
\begin{tabular}{@{}llccc@{}}
\toprule
\textbf{Configuration} & \textbf{Codebook Size} & \textbf{rFID} $\downarrow$ & \textbf{PSNR} $\uparrow$ & \textbf{SSIM} $\uparrow$ \\
\midrule
Semantic Only & 256 & 11.35 & 18.60 & 0.60 \\
Detail Only & 4096 & 5.77 & 19.38 & 0.63 \\
\textbf{Ours} & 256 $\times$ 4096 & \textbf{1.72} & \textbf{20.95} & \textbf{0.67} \\
\bottomrule
\end{tabular}}
\end{table}

\begin{figure}[t]
\centering
\includegraphics[width=0.48\textwidth]{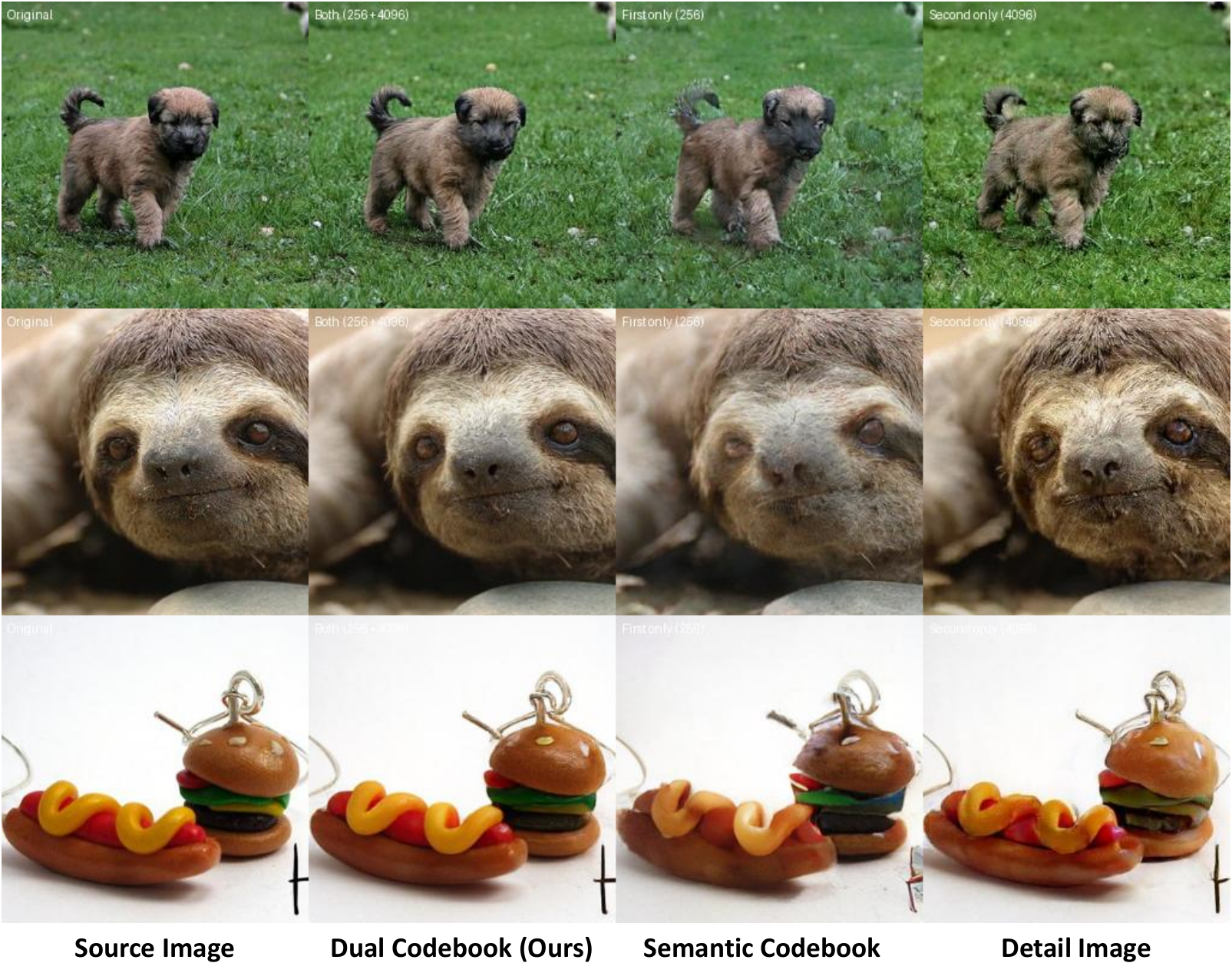} 
\caption{\hutnew{Visual analysis of the dual codebook reconstruction, where both the single codebook (semantic and detail) cannot reconstruct the images well.}}
\label{fig:supp_codebook_vis}
\end{figure}

As visually demonstrated in Figure~\ref{fig:supp_codebook_vis}, decoding with \textit{only} the semantic or detail tokens leads to inadequate image reconstructions. It is only when both token types are combined that fine-grained local textures, sharp edges, and intricate patterns are fully restored. Therefore, employing both the semantic and detail codebooks is essential for achieving a robust and highly expressive image representation.

\subsection{\hutnew{Ablation on Codebook Coupling and Usage}}
In Section 4.2 of the main text, we emphasized that the semantic and detail codebooks are deliberately \textbf{coupled} (associated). Specifically, the detail codebook quantizes the \textit{residual} of the semantic codebook, and during generation, the AR head predicts the detail token \textit{conditioned} on the predicted semantic token. 

To comprehensively evaluate the necessity of both codebooks and their coupling mechanism, we conduct a detailed ablation study investigating the effects of discarding or decoupling the codebooks. The quantitative results are presented in Table~\ref{tab:supp_codebook_ablation}. Note that all models are evaluated at \textbf{100 epochs} to ensure a fair comparison of convergence efficiency. The results can be analyzed as follows:

\begin{table}[h]
\centering
\caption{\hutnew{Ablation study on codebook usage and architecture during generation (at 100 epochs). Discarding either the semantic or detail codebook from our framework severely degrades generation quality. Our proposed semantic-detail association is crucial for effectively leveraging a dual-codebook setup and outperforms all single-codebook and unassociated baselines. Best results are in \textbf{bold}.}}
\renewcommand{\arraystretch}{1.2}
\resizebox{0.48\textwidth}{!}{ 
\label{tab:supp_codebook_ablation}
\begin{tabular}{@{}lcccc@{}}
\toprule
\textbf{Method} & \textbf{FID} $\downarrow$ & \textbf{IS} $\uparrow$ & \textbf{Precision} $\uparrow$ & \textbf{Recall} $\uparrow$ \\
\midrule
Only Semantic Codebook & 9.26 & 179.2 & 0.761 & 0.342 \\
Only Detail Codebook & 11.72 & 180.2 & 0.780 & 0.317 \\
\midrule
Single Codebook  & 6.60 & \textbf{187.2} & \textbf{0.849} & 0.400 \\
\yrexten{Unassociated} Dual Codebook~\cite{song2025dualtoken} & 6.74 & 169.2 & 0.820 & \textbf{0.410} \\
\textbf{Associated Dual Codebook (Ours)} & \textbf{6.29} & 186.3 & 0.840 & \textbf{0.410} \\
\bottomrule
\end{tabular}}
\end{table}

\begin{itemize}
    \item \textbf{Discarding One of the Dual Codebooks:} We first investigate the impact of maintaining the dual-codebook architecture but utilizing only one during the generation phase. If we discard the semantic tokens entirely (using \textit{only} detail tokens), the FID drastically deteriorates to 11.72. Conversely, discarding the detail tokens (using \textit{only} semantic tokens) results in an FID of 9.26. Relying solely on semantic tokens fails to capture fine-grained textures, leading to a poor FID and a drop in Recall, which indicates a severe lack of realistic diversity. This demonstrates that neither codebook can be discarded; they are inherently complementary in the generation process.
    
    \item \textbf{Collapsing to a Single Codebook:} If we discard the semantic-detail division from the beginning and train the model relying on only one standard codebook, it degenerates into the conventional ``Single Codebook'' baseline. This achieves an FID of 6.60. It suffers from the inherent trade-off between reconstruction fidelity and generation quality, as a limited linear vocabulary size cannot perfectly accommodate both global semantics and local details.
    
    \item \textbf{Decoupling the Codebooks (Unassociated Dual Codebook):} If we employ two codebooks but remove their dependency—meaning we do not use the residual formulation and do not condition the detail prediction on the semantic token—the performance degrades to an FID of 6.74. Without the semantic token serving as an anchor, the detail token loses its reference context, making it extremely difficult for the AR model to independently predict unassociated high-frequency information.
    
    \item \textbf{Associated Dual Codebook (Ours):} By strongly coupling the two codebooks, the detail tokens explicitly learn meaningful residuals based on the established semantic context. This hierarchical dependency allows our model to effectively exploit the $n_1 \times n_2$ polynomial capacity, achieving the best overall generation quality with an FID of \textbf{6.29}.
\end{itemize}

In conclusion, neither the semantic nor the detail codebook can be discarded without significantly degrading generation quality. Furthermore, the two codebooks cannot be simply merged into a single codebook or decoupled; they must remain mathematically and conditionally associated to effectively generate coherent visual details.

\section{More Visualization Resutls}
\label{sec:more visualization results}

\begin{figure*}[t]
\centering
\includegraphics[width=1.0\textwidth]{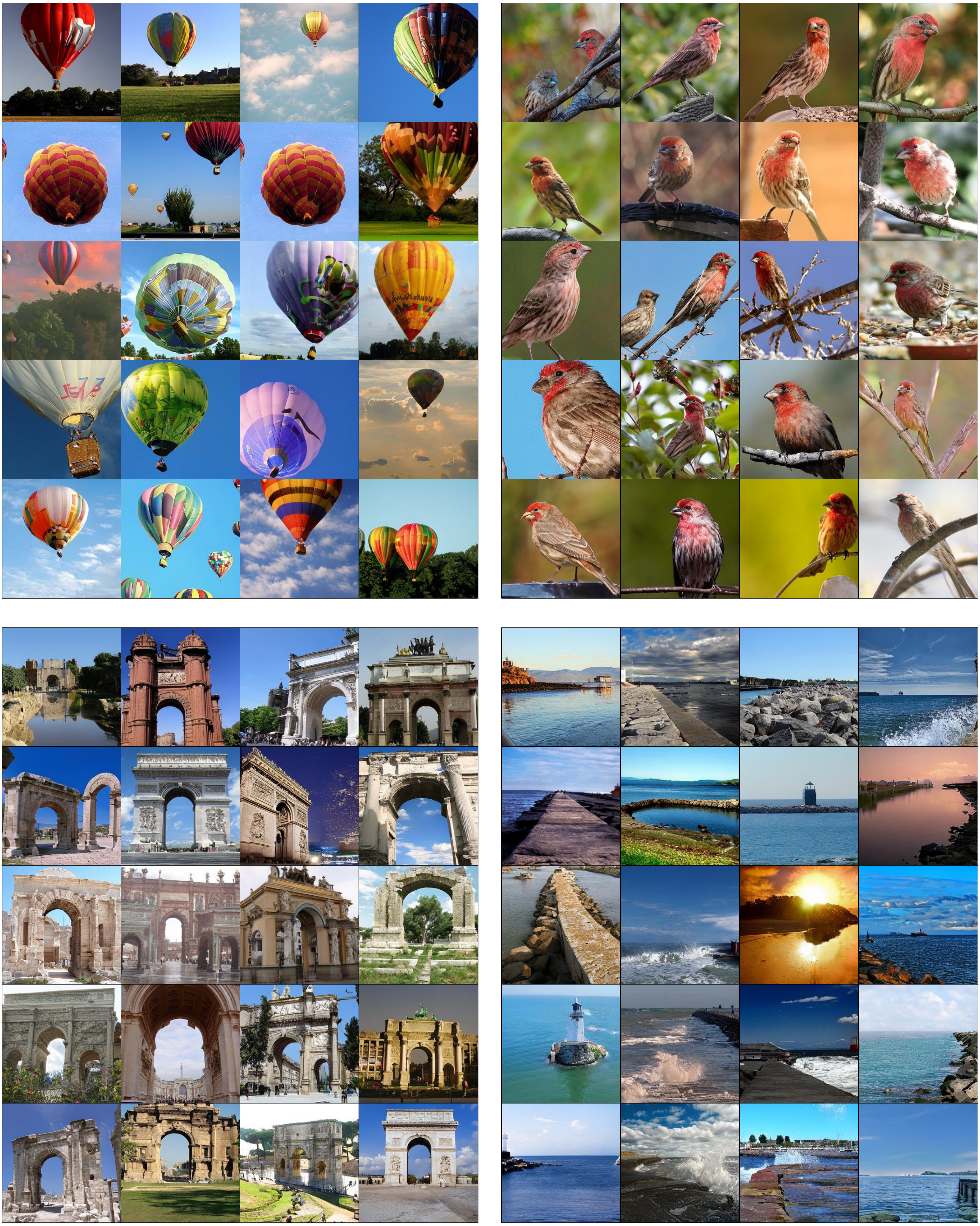}
\vspace{-0.1in}
\caption{The generated images for balloon, house finch, triumphal arch, and breakwater by IAR2-XL.}
\label{fig: generated_image1}
\vspace{-0.15in}
\end{figure*}

\begin{figure*}[t]
\centering
\includegraphics[width=1.0\textwidth]{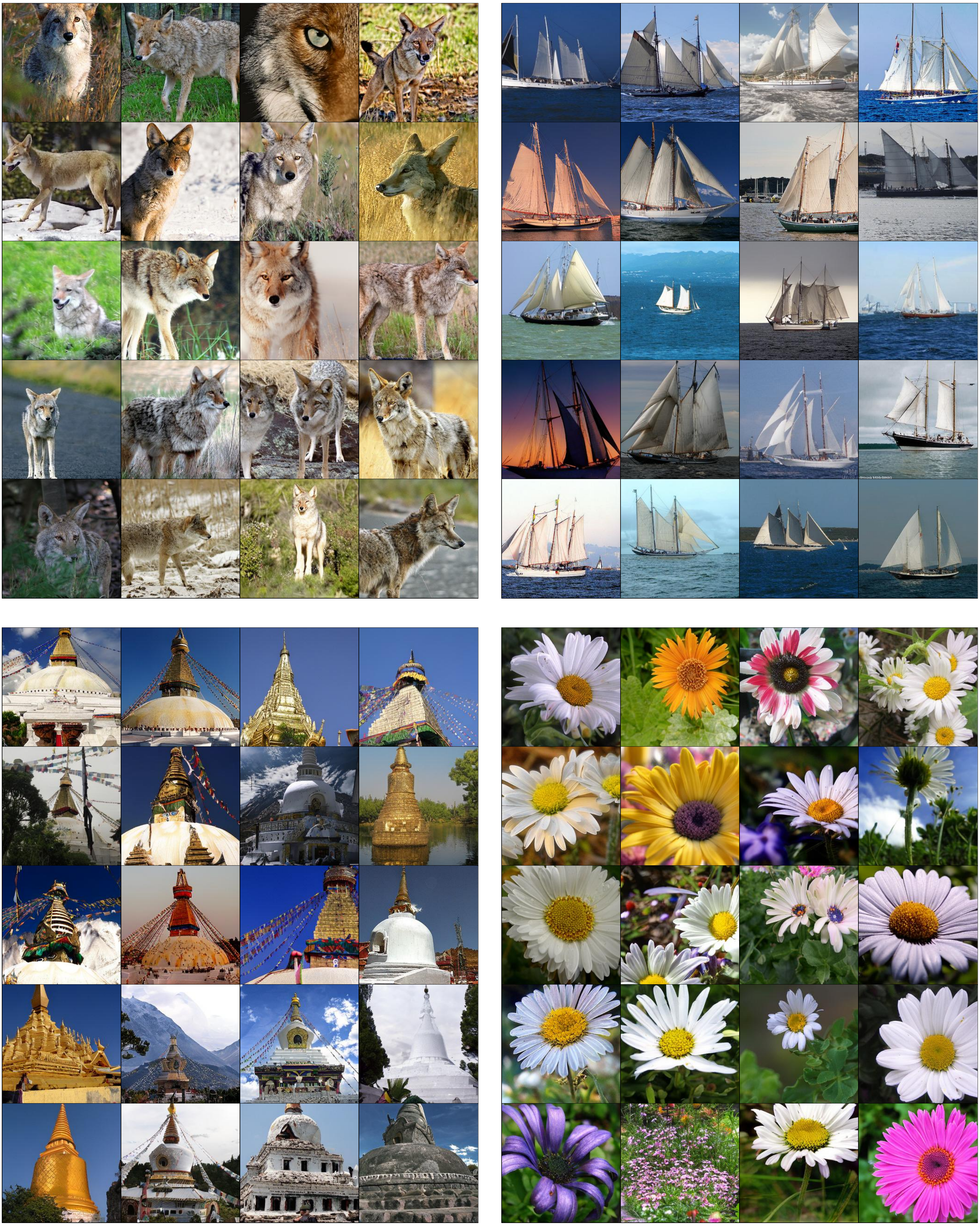}
\vspace{-0.1in}
\caption{The generated images for coyote, schooner, stupa, and daisy by IAR2-XL.}
\label{fig: generated_image3}
\vspace{-0.15in}
\end{figure*}

We \yrexten{show} more generated images from our model in Fig.~\ref{fig: generated_image1}$\sim$\ref{fig: generated_image3}, where the images are generated by the IAR2-XL version with \yrexten{progressive} CFG starting from 1.35 to 3.0, with image size $384\times 384$. We show 12 classes of images, including balloon, house finch, triumphal arch, breakwater, alp, Arctic fox, marmot, liner, coyote, schooner, stupa, and dais.

\begin{figure*}[t]
\centering
\includegraphics[width=1.0\textwidth]{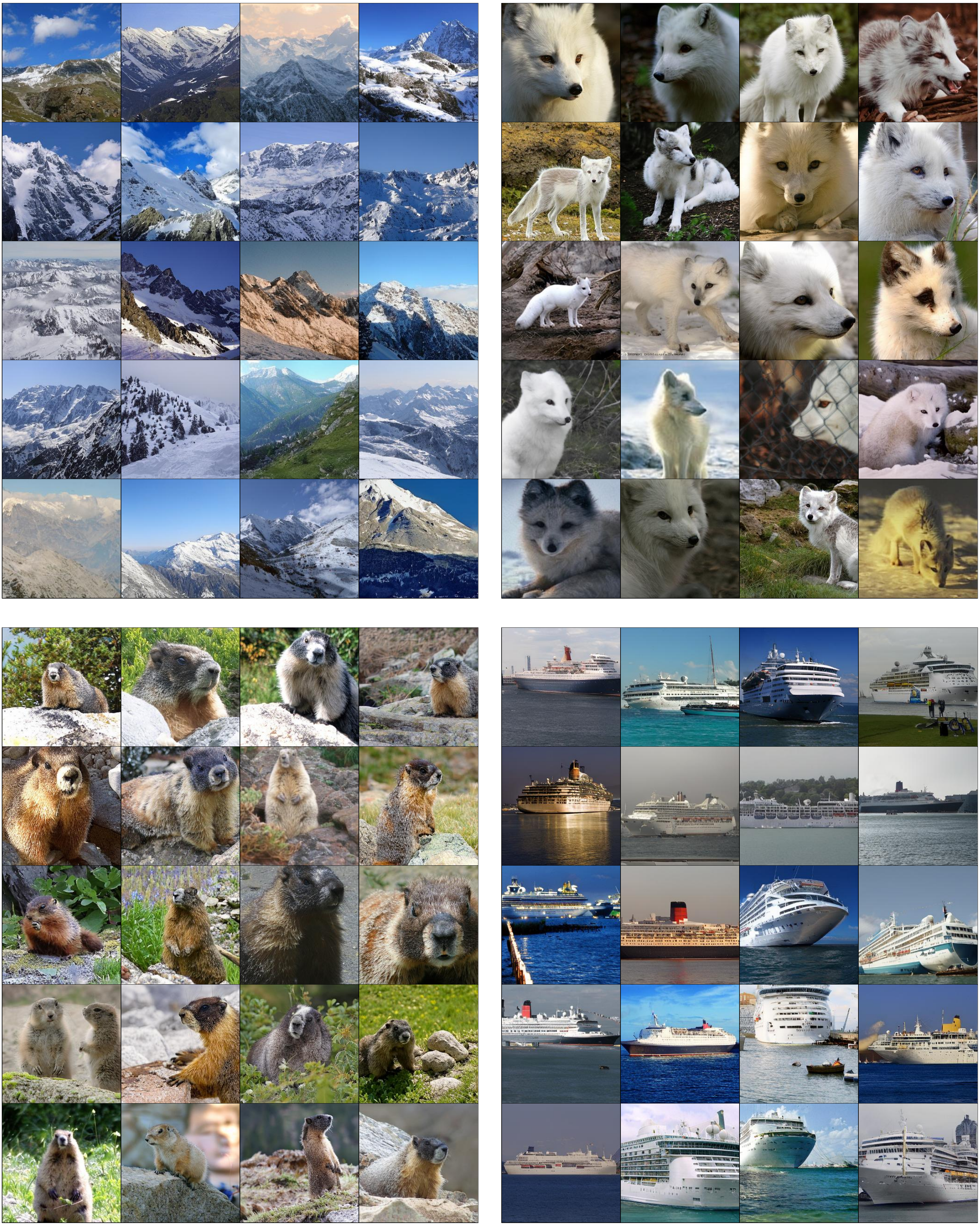}
\vspace{-0.1in}
\caption{The generated images for alp, Arctic fox, marmot, and liner by IAR2-XL.}
\label{fig: generated_image2}
\vspace{-0.15in}
\end{figure*}

\end{document}